\documentclass[10pt,twocolumn,letterpaper]{article}

\usepackage[pagenumbers]{cvpr} 

\usepackage{multirow}
\usepackage{array}
\usepackage{makecell}   
\usepackage{graphicx}
\usepackage{amsfonts}   
\usepackage{amsmath}    
\usepackage{booktabs}   
\usepackage{bm}    
\usepackage{color}      
\usepackage{algorithm}
\usepackage{algorithmic}
\usepackage{setspace}

\usepackage{tikz}
\usepackage{pgfplots}
\usepackage{amssymb}
\usepackage{adjustbox}
\usepackage{bbm}

\usepackage{stfloats} 

\definecolor{cvprblue}{rgb}{0.21,0.49,0.74}
\usepackage[pagebackref,breaklinks,colorlinks,allcolors=cvprblue]{hyperref}

\def\paperID{3737} 
\def\confName{CVPR}
\def\confYear{2025}

\title{SPC-GS: Gaussian Splatting with Semantic-Prompt Consistency for Indoor Open-World Free-view Synthesis from Sparse Inputs}

\author{Guibiao Liao$^{1,2}$ \quad\quad  Qing Li$^{2}$ \quad\quad  Zhenyu Bao$^{1,2}$ \quad\quad  Guoping Qiu$^{3}$ \quad\quad  Kanglin Liu$^{2}$\footnotemark[1]  \\ 
\textsuperscript{\rm 1}School of Electronic and Computer Engineering, Peking University \\ 
\textsuperscript{\rm 2}Pengcheng Laboratory, 
\quad \textsuperscript{\rm 3}University of Nottingham 
}

\begin{document}

\twocolumn[{
\renewcommand\twocolumn[1][]{#1}
\maketitle
\begin{center}
\centering
\includegraphics[width=.99\linewidth]{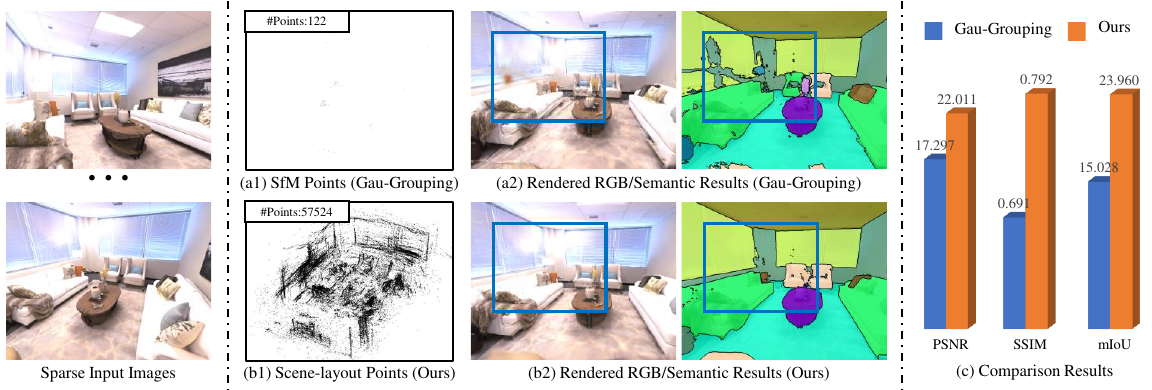}
\vspace{-2mm}
\captionof{figure}{
Visual comparisons of open-world free-view synthesis from sparse inputs (12 training views). 
The previous 3DGS-based method, Gau-Grouping \cite{gaugrouping}, utilizes sparse SfM points for Gaussian initialization and limited training views for supervision, leading to inferior rendering results. 
In contrast, our approach leverages scene-layout-based points for instructive initialization, and cooperates with semantic-prompt consistency constraints (detailed in Sec. \ref{SPC_design}), yielding superior reconstruction and segmentation results. 
} 
\label{fig:vis_fig1}
\end{center}
}]

\renewcommand{\thefootnote}{} 
\footnotetext{$\ast$ Corresponding author}  


\begin{abstract}

3D Gaussian Splatting-based indoor open-world free-view synthesis approaches have shown significant performance with dense input images. However, they exhibit poor performance when confronted with sparse inputs, primarily due to the sparse distribution of Gaussian points and insufficient view supervision. 
To relieve these challenges, we propose SPC-GS, leveraging Scene-layout-based Gaussian Initialization (SGI) and Semantic-Prompt Consistency (SPC) Regularization for open-world free view synthesis with sparse inputs. 
Specifically, SGI provides a dense, scene-layout-based Gaussian distribution by utilizing view-changed images generated from the video generation model and view-constraint Gaussian points densification. 
Additionally, SPC mitigates limited view supervision by employing semantic-prompt-based consistency constraints developed by SAM2. 
This approach leverages available semantics from training views, serving as instructive prompts, to optimize visually overlapping regions in novel views with 2D and 3D consistency constraints. 
Extensive experiments demonstrate the superior performance of SPC-GS across Replica and ScanNet benchmarks. Notably, our SPC-GS achieves a 3.06 dB gain in PSNR for reconstruction quality and a 7.3\% improvement in mIoU for open-world semantic segmentation. 
Project website at: \url{https://gbliao.github.io/SPC-GS.github.io}. 

\end{abstract}

\section{Introduction} \label{sec:intro} 
3D Gaussian Splatting-based open-world free-view synthesis seeks to construct semantic Gaussian radiance fields using a set of 3D Gaussian primitives enriched with color and semantic attributes. 
This task enables the free novel view synthesis of both realistic appearances and open-vocabulary semantic understanding (\textit{also referred to as 3D open-vocabulary semantic segmentation}) of the scenes, with significant implications for augmented/virtual reality, robot navigation, and manipulation \cite{zhang2022tdrnet, robotnavigation, liao2024vlm2scene, conceptfusion, liao2020mmnet, liu2024moka}. 
Existing methods \cite{feature3dgs,langsplat,gaugrouping} generally require dense, large sets of input views to train the semantic Gaussian radiance field for comprehensive scene coverage. However, when only sparse views are available, these approaches exhibit substantial declines in reconstruction and segmentation accuracy. This is mainly attributed to two key limitations.

\textbf{1) Sparsely Distributed Gaussian Points}: 
With sparse input views, the initialized Gaussian points via Structure-from-Motion (SfM) in 3DGS \cite{3dgs} are inherently few and sparsely distributed (Fig. \ref{fig:vis_fig1} (a1)). Such sparse point distribution imposes challenges in optimizing 3D Gaussians to depict complex indoor environments, leading to suboptimal reconstruction quality and, consequently, impaired segmentation accuracy (Fig. \ref{fig:vis_fig1} (a2)). 
\textbf{2) Insufficient View Constraints}: 
Optimizing 3D Gaussian parameters with inadequate view supervision often converges to local minima that overfit training views \cite{dngaussian,fsgs}. This results in semantic ambiguity and poor performance in under-observed regions when rendering from novel viewpoints, due to limited reliable semantic integration across different views.

To address the above challenges, we propose \textbf{SPC-GS}, a novel framework that incorporates \textbf{Scene-layout-based Gaussian Initialization (SGI)} and \textbf{Semantic-Prompt Consistency (SPC)} regularization.

To seek denser points for improved Gaussian initialization, SGI produces a structured scene-layout Gaussian distribution in two steps. 
{First}, SGI employs a video generation model to create view-changed images proximal to training images. These generated images, combined with training views, enrich feature matching in SfM, yielding denser SfM points that promote scene-layout Gaussian generation. 
{Second}, these denser SfM points undergo view-constraint Gaussian densification and outlier removal to generate structured scene-layout points. 
These points provide an instructive initialization that faithfully captures indoor scene geometry (\eg furniture, walls, ceilings), enhancing 3DGS training and scene representation, thereby improving segmentation accuracy (Fig. \ref{fig:vis_fig1} (b1) and (b2)).

Building on dense scene-layout Gaussian initialization, SPC enhances supervision through view-to-view semantic consistency constraints. 
The core of SPC lies in utilizing informative semantics signals from known training views, serving as \textit{semantic prompts}, to optimize online-rendered novel (pseudo) views, establishing pseudo-multi-view constraints through training-to-pseudo semantic consistency. 
To achieve it, we utilize the powerful Segment Anything Model 2 (SAM2) \cite{sam2} to build region mask correspondences across training and pseudo views.  
Using these correspondences, we leverage the semantics prompts provided by training views to enforce semantic consistency constraints on the corresponding regions of pseudo views at 2D and 3D spaces. 
This approach substantially augments supervision signals, leading to more coherent segmentation on novel views under sparse input conditions.

Prior works on 3D open-vocabulary segmentation or sparse-input free-view synthesis mainly focus on \textit{object-centric} or \textit{face-forward} scenarios, such as LERF \cite{LERF}, and 3DOVS \cite{3DOVS}, where objects are centrally positioned, and views are sampled in an ``outside-in" manner, resulting in significant view overlap. 
In contrast, this study focuses on sparse-input open-world free-view synthesis within \textit{indoor scenes}, particularly relevant for robotics and augmented/virtual reality applications. 
To this end, we utilize the widely recognized Replica \cite{replica} and ScanNet \cite{scannet} datasets, which feature representative indoor scenes characterized by an ``inside-out" viewing configuration. This setup leads to less overlap between certain views, posing a substantial challenge for sparse-input open-world free-view synthesis. 
The contributions of this work are as follows:

\begin{itemize} 
    \item We propose SPC-GS, an early and novel attempt at jointly reconstructing and understanding open-world indoor scenes from sparse-input images. 
    \item We introduce Scene-layout-based Gaussian Initialization (SGI) to tackle the challenge of sparsely distributed Gaussians, offering a dense and instructive scene-layout point initialization that promotes semantic Gaussian learning. 
    \item We present Semantic-Prompt Consistency (SPC) regularization to alleviate inadequate view supervision by enforcing semantic consistency constraints on unseen pseudo views through prompts derived from training views, enhancing coherent segmentation.  
    \item Extensive experiments on sparse-input scenes show the superiority of our approach in free-view synthesis of realistic appearances and open-vocabulary segmentation. 
\end{itemize}

\section{Related Work}
\textbf{Free-view Synthesis via Radiance Fields.} 
In recent years, radiance field-based free-view synthesis has made remarkable advancements. 
A Notable example is the Neural Radiance Field (NeRF) \cite{nerf, nerfreview}, which leverages a coordinate-based multi-layer perceptron (MLP) to implicitly model 3D scenes, enabling novel view synthesis via volume rendering \cite{drebin1988volume}. 
Another innovative approach is 3D Gaussian Splatting (3DGS) \cite{3dgs, wu2024recent}, which utilizes a collection of 3D Gaussians to represent the scene and incorporates a fast tile-based rasterizer technique, allowing real-time rendering. 

Although current methods achieve impressive results in dense-view scenarios \cite{instantngp,tensorf,2dgs,ovnerf,mipsplatting,liao2024clip,fei2024recent}, their performance degrades significantly with sparse inputs \cite{dirtnerf, fsgs}.

\noindent \textbf{Sparse-input Free-view Synthesis. }
Recent efforts have tried to address the sparse-input view synthesis and they can broadly categorized into cross-scene pre-training methods and per-scene optimization methods.  

\textit{Pre-training methods} learn scene priors by pre-training on large-scale datasets spanning multiple scenes. 
For instance, pixelNeRF \cite{pixelnerf} conditions a NeRF on CNN features from input images during volume rendering. Similarly, several approaches \cite{ibrnet, mvsnerf} utilize different NeRF conditions to facilitate cross-scene pre-training. 
Inspired by the success of 3DGS, recent methods such as pixelSplat \cite{pixelsplat} and MVSplat \cite{mvsplat} integrate efficient CUDA-based rasterization with 3D Gaussian representations for pre-training. However, these works demand substantial training resources and large-scale datasets. 

\textit{Per-scene optimization methods} aim to rapidly reconstruct a 3D scene by introducing effective regularizations. 
Early NeRF-based methods enhance sparse-input view synthesis through depth-based \cite{regnerf, sparsenerf} and frequency range regularization \cite{freenerf}. 
Recent 3DGS-based works focus on improving rendering efficiency. 
\cite{dr3dgs} designs depth regularization loss using monocular depth. 
DNGaussian \cite{dngaussian} and FSGS \cite{fsgs} enhance sparse Gaussian representations through global-local depth normalization and correlation depth distribution losses. 
CoR-GS \cite{CoRGS} adopts color co-regularization loss using rendered pseudo views. 

Instead, we focus on enhancing view synthesis and open-world 3D indoor semantic understanding from sparse inputs following the per-scene optimization route.

\noindent \textbf{3D Open-world Understanding via Radiance Fields.}  
Recent studies have explored integrating 2D visual foundation models (\eg DINO \cite{DINO} and CLIP \cite{OpenAICLIP, LSeg}) with radiance fields to facilitate 3D open-world scene understanding. These approaches can be broadly divided into NeRF-based and 3DGS-based methods.

\textit{NeRF-based}. 
Early methods \cite{n3f, nerfsos} distill visual features from DINO into semantic radiance fields. 
More recent language-embedded neural field methods \cite{DFF, LERF} integrate CLIP's visual features into semantic fields, enabling language-driven 3D semantic understanding. 
Additionally, 3DOVS \cite{3DOVS} further improves semantic accuracy through the proposed Relevancy-Distribution Alignment loss.

\textit{3DGS-based.} 
Alternative to NeRFs, recent works explore embedding CLIP features into 3D Gaussians for more efficient semantic rendering.  
Feature 3DGS \cite{feature3dgs} directly attaches CLIP embeddings to 3D Gaussians, while FMGS \cite{fmgs} combines Gaussians with multi-resolution hash encodings to form the semantic Gaussian field. 
\cite{LEGaussian} embeds quantized CLIP features into Gaussians to reduce storage demand.
Furthermore, LangSplat \cite{langsplat} learns low-dimensional semantic Gaussian embeddings and uses a deep neural network for post-process upsampling to obtain semantic representation. 
Gau-Grouping \cite{gaugrouping} introduces a 2D object identity loss within training views to optimize Gaussians, while this method requires dense view supervision. 
Note that the 3D interactive segmentation methods, such as \cite{saga,flashsplat,kim2024garfield,ying2024omniseg3d}, lie outside the scope of language-driven open-world segmentation, as they rely on 2D point prompts to generate object masks but lack semantic meaning. 

While previous approaches rely on dense input images, our work focuses on GS-based 3D open-world scene understanding in sparse-input indoor scenarios.

\begin{figure*}
\centering
\includegraphics[width=\linewidth]{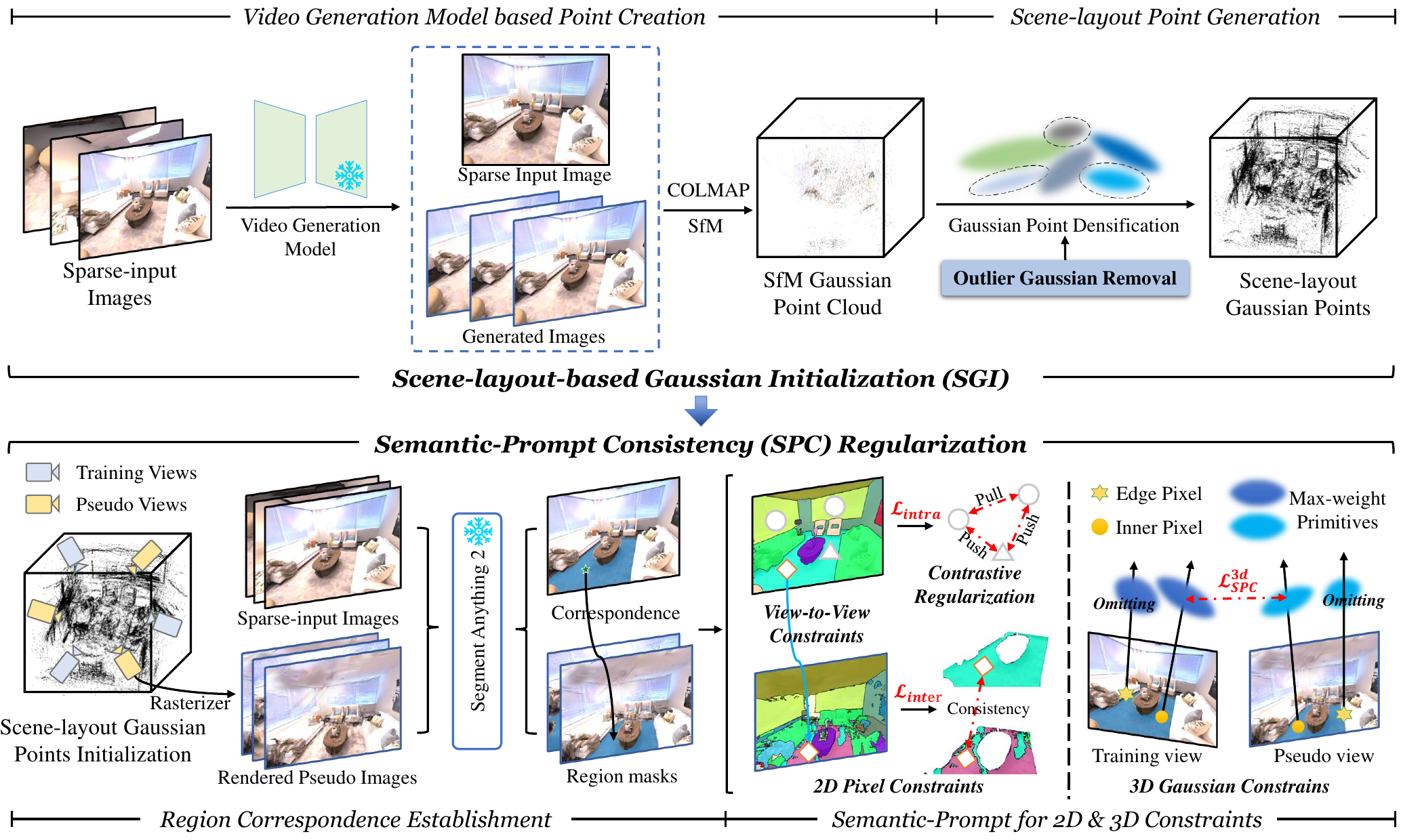}
\vspace{-6mm}
\caption{
\textbf{Framework of SPC-GS.} 
(a) We first generate adjacent images of each training image using the video generation model \cite{motionctrl}. These generated images, combined with the original training images, produce denser initialized SfM points. These points are then optimized to create a scene-layout Gaussian distribution via Gaussian densification and outlier removal. 
(b) Building on the scene-layout Gaussian initialization, SPC leverages semantic information from training views as instructive semantic prompts to optimize adjacent rendered pseudo views, establishing semantic consistency constraints that enhance overall sparse-input semantic understanding of 3D scenes. 
}
\label{fig:overview}
\end{figure*}

\section{Methodology}
Fig. \ref{fig:overview} illustrates our framework, which consists of two primary components. 
First, Scene-layout-based Gaussian Initialization (SGI) (Section \ref{SGI_design}) constructs a dense, instructive scene-layout Gaussian initialization that improves Gaussian representation, thus supporting enhanced semantic learning. 
Then, using the initialized scene-layout Gaussians, Semantic-Prompt Consistency (SPC) Regularization (Section \ref{SPC_design}) introduces 2D and 3D semantic consistency constraints, creating effective pseudo-multi-view optimization for semantic learning under sparse-input conditions.

\subsection{Preliminary} \label{Overview_framework} 
\textbf{Problem Definition.}  
Our approach aims to leverage sparse input images to reconstruct 3D indoor scenes and enable language-driven semantic perception. 
Concretely, given $L$ posed sparse input images ${I} = \{ {I}_i \}_{i=1}^{L} \in \mathbb{R}^{L \times H \times W \times 3} $ and language descriptions, our goal is to generate both realistic appearance and open-world semantic understanding results for specified novel viewpoints.  
To achieve this, we exploit 3D Gaussians \cite{3dgs} for scene representation, and utilize CLIP models \cite{OpenAICLIP, LSeg} to provide semantic supervision.

\noindent \textbf{Preliminary.} 
3D Gaussian Splatting (3DGS) \cite{3dgs} represents a 3D scene using a collection of explicit 3D Gaussian primitives $\{ \mathbf{g}_{i} = (\mathbf{p}_{i}, \mathbf{c}_{i}, o_{i}, \mathbf{s}_{i}, \mathbf{q}_{i} ) \}_{i=1}^{N_K}$, where each primitive comprises a 3D position $\mathbf{p} \in \mathbb{R}^3$, color $\mathbf{c} \in \mathbb{R}^3$, opacity $o \in \mathbb{R}$, scaling factor $\mathbf{s} \in \mathbb{R}^3$, and rotation quaternion $\mathbf{q} \in \mathbb{R}^4$. 
For pixel color $\hat{I}(x_{p})$ rendering, 3DGS employs $\alpha$-blending point-based rendering \cite{kopanas2021point}, formulated as:
\begin{equation}
\hat{I}(x_{p}) = \sum_{i \in \mathcal{N}} T_i \alpha_i \mathbf{c}_i,  
\end{equation} 
where \(\mathcal{N}\) represents Gaussian points overlapping pixel $x_{p}$, $\alpha_{i} = o_i G_i^{proj}$ with $G_i^{proj}$ as the $i$-th Gaussian's 2D projection, and transmittance $T_i = \prod_{j=1}^{i-1} (1 - \alpha_j)$.

For pixel semantic feature $\hat{\mathbf{F}}(x_{p})$ rendering, following \cite{feature3dgs,langsplat}, we attach a learnable semantic parameter $\mathbf{f}_{i} \in \mathbb{R}^D$ to each Gaussian and use the above $\alpha$-blending: 
\begin{equation} 
\hat{\mathbf{F}}(x_{p}) = \sum_{i \in \mathcal{N}} T_i \alpha_i \mathbf{f}_i. 
\label{eq_alphablending_sem}
\end{equation} 
Following \cite{3DOVS}, we use a CLIP text encoder to generate textual feature $\mathbf{T} \in \mathbb{R}^{M \times 512}$ for $M$ classes and obtain rendered segmentation logits $\hat{\mathbf{S}}$ as follows: 
$\hat{\mathbf{S}} = \mathrm{cos} \langle \omega_{f}(\hat{\mathbf{F}}), \mathbf{T} \rangle$, where $\mathrm{cos}$ denotes cosine similarity and $\omega_{f}$ is a 1 $\times$ 1 convolution layer for feature dimension alignment.

We incorporate reconstruction and semantic losses to train the 3D Gaussians.  
The color reconstruction loss, denoted as $\mathcal{L}_C$, follows the formulation in 3DGS \cite{3dgs}, and the semantic loss, $\mathcal{L}_S$, is defined as: 
\begin{equation}
\mathcal{L}_S = \mathcal{L}_{cos}( \omega_{f}(\hat{{\mathbf{F}}}), {{\mathbf{F}}} ) + \mathcal{L}_{ce}( \omega_{s}(\hat{\mathbf{S}}), {\mathbf{S}} ), 
\label{semantic_loss}  
\end{equation} 
where ${\mathbf{F}}$ and ${\mathbf{S}}$ are CLIP-derived semantic features and relevancy maps, respectively. $\omega_s$ is a 1 $\times$ 1 convolution layer, and $\mathcal{L}_{cos}$ and $\mathcal{L}_{ce}$ denote cosine and cross-entropy losses.

\subsection{Scene-layout-based Gaussian Initialization}  \label{SGI_design}
We propose Scene-layout-based Gaussian Initialization (SGI) to produce a dense and conducive point distribution for initialization with the following insights and designs.

\noindent
\textbf{VGM-based Point Creation.}  
In static indoor scenes, viewpoint variations correspond to camera movements. 
Motivated by this, we employ a video generation model (VGM) to create additional views beyond the sparse training set through image-to-video generation. These generated views augment the available scene information and, when combined with the original training views, enrich feature matching during SfM, yielding denser SfM points for generating scene-layout Gaussians.

Specifically, for each training image $I_i$, we generate eight adjacent views $\{\tilde{I}_i^j\}_{j=1}^{8}$ using the image-to-video generation mode of the powerful video generation model MotionCtrl \cite{motionctrl}. 
This is accomplished by inputting the training images along with eight standard camera motion trajectories \cite{motionctrl} into MotionCtrl. 
These trajectories include \textit{pan up, pan down, pan left, pan right, zoom in, zoom out, clockwise rotation}, and \textit{anticlockwise rotation}. 
Each trajectory generates a sequence of consecutive video frames. 
From each sequence, we select a single frame temporally adjacent to the input training image as the valid generated image since it exhibits superior visual quality. 
These generated images are then combined with the original training images for SfM processing, yielding denser SfM points for Gaussian initialization (\textit{see supplementary file}). 
Despite being denser than SfM points derived from sparse-input training images, these points remain limited in capturing the geometric layout and details of indoor scenes.

\noindent
\textbf{Scene-layout Point Generation.}
We further generate scene-layout point distributions to enhance 3DGS training and scene representation through the following steps.

\textit{i) View-constraints 3D Gaussian Densification.}
Using the denser SfM point initialization, we follow Gaussian densification \cite{3dgs} and using our view constraints from Eq. \eqref{overall_loss} to generate a more densely distributed, instructive scene-layout Gaussian point set $\{ \mathbf{g}_{i} = (\mathbf{p}_{i}, \mathbf{c}_{i}, \mathbf{f}_i, o_{i}, \mathbf{s}_{i}, \mathbf{q}_{i} ) \}_{i=1}^{N_{K'}}$, as shown in Fig. \ref{fig:overview}. 
These denser $N_{K'}$ points provide an improved spatial layout of indoor scenes, and enhanced attribute representations (\eg $\mathbf{c}_{i}$, $\mathbf{f}_i$), making them effective priors for Gaussian initialization.

\textit{ii) Outlier Gaussian Primitive Removal (OGR).}  
During Gaussian densification, new Gaussian primitives are sampled based on a 3D Gaussian probability distribution, which may generate outlier primitives positioned far from the actual scene layout. These outliers, when used in subsequent initialization, complicate the optimization process. 
To mitigate this, we present a density-based removal strategy to remove those sparse outlier points. 
Specifically, Gaussian primitives with fewer than $\mathcal{P}$ neighboring points within a spherical radius $r$ are periodically eliminated during training. 
Empirical experiments show that setting $\mathcal{P}$ to 5 and $r$ to 1 strikes a balance between effective outlier removal and the preservation of essential scene elements.

By using instructive scene-layout-based points for initialization, the 3DGS training and scene representation can be enhanced, consequently improving segmentation performance under the sparse-input condition.

\subsection{Semantic-Prompt Consistency Regularization} \label{SPC_design} 
We introduce Semantic-Prompt Consistency (SPC) regularization to exploit additional supervision signals. The key idea of SPC is to leverage semantics from training views as prompts, to impose effective semantic consistency constraints on corresponding regions of pseudo views.

\noindent
\textbf{Region Correspondence Establishment.}  
To establish correspondences between training and pseudo views, we draw inspiration from the powerful zero-shot temporal region proposal extraction capability of Segment Anything Model 2 (SAM2) \cite{sam2} and leverage it to achieve this.

\begin{figure}[!t]
\centering
\includegraphics[width=.99\linewidth]{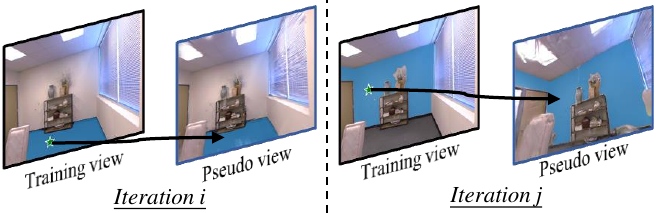}
\vspace{-3mm}
\caption{
Illustration of Iterative Stochastic Prompting (ISP). During training, ISP randomly samples point coordinates ($\bigstar$) from the training view, serving as the point prompt for SAM2 to produce region masks (highlighted in blue) across training and pseudo views. 
}
\label{fig:vis_ISP}
\end{figure}

Specifically, we introduce Iterative Stochastic Prompting (ISP) to efficiently build correspondence. During training, ISP renders pseudo (unobserved) views using the online training SPC-GS. The view locations are sampled from the two nearest known training views by adding random noise following \cite{fsgs}. 
Then, ISP treats the current training image and its associated pseudo-views as a viewpoint-changed video sequence, and selects a stochastic point coordinate in the training image, as the 2D point prompt for SAM2, generating corresponding region masks across both the training and pseudo views, which can be formulated as: 
\begin{equation}  
\mathbf{M}_i, \{ \mathbf{M}^{*}_i\}_{p=1}^{P} =  \mathbf{ISP} ({I}_i, \{ {I}_i^{*} \}_{p=1}^{P}), 
\end{equation}  
where ${I}_i^{*}$ is the rendered pseudo image and $\mathbf{M}^{*}_i$ is the corresponding region mask within ${I}_i^{*}$. $P$ is set to 2 for computational efficiency. 
As training progress proceeds, the stochastic sampling mechanism encourages a uniform distribution of points across the entire image space, facilitating the efficient region mask correspondence construction for diverse regions of the whole image, as shown in Fig. \ref{fig:vis_ISP}.

\noindent
\textbf{Semantic-Prompt for 2D Constrains. }
Building on the established region mask correspondences, we enforce effective constraints within 2D views to enhance Gaussian learning with sparse inputs. 

\textit{i) View-to-View Regularization. }
Leveraging the semantics from training views as prompts, we enforce training-to-pseudo semantic consistency constraints in feature and relevancy map aspects, which is formulated as: 
\begin{align}
    \mathcal{L}_{inter} = & \sum_{p = 1}^{P} 
    || \frac{ \hat{\mathbf{F}}^{*}_{p} \odot \mathbf{M}^{*}_{p} }{\sum {\mathbf{M}^{*}_{p}}} 
    - \frac{ \hat{\mathbf{F}} \odot \mathbf{M} }{\sum {\mathbf{M}} }  ||_{2}
    \notag 
    \\ & - \sum_{p = 1}^{P} \mathcal{R} [\Phi(\hat{\mathbf{S}} \odot \mathbf{M}) ] \mathrm{log} ( \hat{\mathbf{S}}^{*}_{p} \odot \mathbf{M}^{*}_{p} ), 
    \label{eq_interloss}
\end{align}
where $\hat{\mathbf{F}}^{*}$ and $\hat{\mathbf{S}}^{*}$ represent the semantic feature and the relevancy map derived from pseudo views. 
The function $\Phi$ is the semantic uniformity operation: $\Phi(\hat{\mathbf{S}} \odot \mathbf{M}) = {\mathbf{argmax}}_{c} \sum_{i \in \mathbf{M}} \mathbbm{1} \{ \mathbf{argmax}(\omega_{s} (\hat{\mathbf{S}}_{i}) ) = c \}$, which integrates SAM2's masks with a maximum voting mechanism to select the dominant class $c$ within the region mask $\mathbf{M}$, unifying the region’s semantics and enforcing consistent region-wise semantic supervision.
$\mathcal{R}[\cdot]$ denotes the dimensional reshaping operation to align the dimension shape of the region mask $\mathbf{M}^{*}$ from the pseudo view. 
In $\mathcal{L}_{inter}$, we utilize self-rendered semantic ($\hat{\mathbf{F}},\hat{\mathbf{S}}$) from training views, generated by the trained SPC-GS, rather than 2D CLIP semantics. This is motivated by the superior coherence and reliability of 3D representations over their 2D counterparts \cite{yue2024improving} 
(\textit{see supplementary file for experimental validation}).

\textit{ii) Contrastive Regularization. } 
Beyond inter-view, we introduce semantic-guided contrastive learning regularization within intra-training-views to further enhance supervision. This is based on the principle that features of semantically consistent objects should be similar.

Given $N$ feature samples, each sample $\mathbf{z}$ is derived from $ \mathbf{z} = W_{\psi}(\hat{\mathbf{F}}) $, where $W_{\psi}$ is a 1 \(\times\) 1 convolution transformation layer. 
Let $i \in U_{z} \equiv \{1...N\}$ be the index of a sample, and $T_{z}(i) \equiv U_{z} \backslash \{i\}$ represents the index set excluding index $i$.
We designate features with the same rendered semantic label $\mathbf{\tilde{q}}$ as positive samples, where \(\mathbf{\tilde{q}} = \Phi(\hat{\mathbf{S}} \odot \{\mathbf{M}^{k}\}_{k=1}^{\eta} ) \). Here, the $\{\mathbf{M}^{k}\}_{k=1}^{\eta}$ denote ${\eta}$ masks generated by SAM2's automatic mask generator in the entire image, and $\Phi$ is the function that unifies semantics within each mask. 
Accordingly, for the $i$-th sample, positive sample indexes are defined as: $P_{z}(i) \equiv \{ p \in T_{z}(i): \mathbf{\tilde{q}}_{p} = \mathbf{\tilde{q}}_{i} \}$, and negative sample indexes: ${ T_{z}(i) \backslash P_{z}(i) }$. 
Similar to SimCLR \cite{SimCLR}, we employ contrastive learning loss to maximize agreement between positive samples. 
This method brings $P_{z}$-indexed positive samples closer in feature space, imposing effective contrastive constraints on sparse-view Gaussian learning, which can be formulated as: 
\begin{equation} 
    \mathcal{L}_{intra} = - \sum_{i \in U_{z}} \frac{1}{\mathcal{M}_{z}(i)} \sum_{p \in P_{z}(i)} \log 
    \left[
    \frac{e^{\left( (\mathbf{z}_i \cdot \mathbf{z}_p) / \tau\right)}}
    {\sum_{j \in T_{z}(i)} e^{\left( (\mathbf{z}_i \cdot \mathbf{z}_j) / \tau\right)}} \right],  
\label{eq_intra_contrastive}
\end{equation}
where $\mathcal{M}_{z}(i)$ denotes the number of elements in $P_{z}(i)$. The symbol $\cdot$ refers to the inner product operation. $\tau$ is the temperature parameter, set to 0.2 in the experiment.

To obtain $N$ samples, we employ a patch-wise sampling strategy by randomly extracting a patch of size $h \times w$ in each training iteration, where $N = h\times w$. Empirically, $h$ and $w$ are both set to 128. 
In summary, the semantic prompt for 2D constraints can be formulated as: 
\begin{equation}
\mathcal{L}_{SPC}^{2d} = \mathcal{L}_{inter} + \mathcal{L}_{intra}. 
\end{equation}

\noindent
\textbf{Semantic-Prompt for 3D Constrains. } 
In addition to 2D constraints, we extend semantic-prompt to impose effective 3D constraints on 3D Gaussian primitives. 
Considering that Gaussian primitives with maximum weight dominate the rendering results (Eq. \eqref{eq_alphablending_sem}), we encourage \textit{semantic feature similarity} between the maximally-weighted 3D Gaussians from two corresponding regions. 
Specifically, for each pixel within the region mask, we identify the Gaussian primitive with maximum weight along the ray cast into 3D space. Then, the semantic features of these maximally-weighted Gaussians from training views are used to constrain their counterparts in pseudo views.

However, since Gaussians related to region boundaries often span objects with differing semantics, directly applying the above constraints may complicate optimization. 
To mitigate this, we introduce a \textit{Region Boundary Erosion} technique, which removes ambiguous primitives corresponding to region boundaries, which is defined as: 
\begin{equation}
\mathcal{L}_{SPC}^{3d} = \mathcal{L}_{kl}( \mathcal{F} || \mathcal{F}^{*} ) = \sum_{j = 1}^{n'} \mathbf{\bar{f}} \mathrm{log}( \frac{\mathbf{\bar{f}}}{ \mathbf{f}^{*}_j } ), 
\end{equation} 
where $\mathcal{F}$ denotes the mean feature representation of 3D Gaussian primitives derived from mask $\mathbf{M}'$ in the training view, defined as $\mathbf{\bar{f}} = (\mathbf{f} \odot \mathbf{M}' ) / \sum \mathbf{M}'$. 
Here, $\mathbf{M}'$ refers to the eroded mask after region boundary erosion, which is achieved through morphological erosion. Specifically, a pixel $\mathbf{M}'(x, y)$ is retained as ``True" only if all pixels within its $5 \times 5$ neighborhood in the input original mask $\mathbf{M}$ are ``True"; otherwise, it is set to ``False". This operation effectively shrinks the boundaries of the region, causing them to contract inward. 
The set $\mathcal{F}^{*} = \{\mathbf{f}^{*}_1, \mathbf{f}^{*}_2, ..., \mathbf{f}^{*}_{n'} \}$ comprises $n'$ Gaussian primitives obtained from eroded masks $\{{\mathbf{M}^{*}}'\}_{p=1}^{P} $ in pseudo views.

To summarize, the overall loss is given by: 
\begin{equation}
\mathcal{L} = \mathcal{L}_{C} + \mathcal{L}_{S} + \mathcal{L}_{GS} + \mathcal{L}_{S}^{3d} + \lambda (\mathcal{L}_{SPC}^{2d} + \mathcal{L}_{SPC}^{3d}) ,
\label{overall_loss}
\end{equation} 
where $\mathcal{L}_{GS}$ is the semantic loss for generated images from video generation model: $\mathcal{L}_{GS} = \mathcal{L}_{cos}( \omega_{f}( \hat{{\mathbf{F}}}^{g} ), {\mathbf{F}}^{g} ) + \mathcal{L}_{ce}( \omega_{s}( \hat{\mathbf{S}}^{g} ), \mathbf{S}^{g} )$. 
These generated RGB images are not used for color constraints due to their error-prone color pixels that degrade overall results (\textit{see supplementary file}). 
$\mathcal{L}_{S}^{3d}$ is the 3D local adaptive regularization, which promotes semantic similarity among neighboring Gaussians: $\mathcal{L}_{S}^{3d} = \frac{1}{\theta k} \sum_{i = 1}^{\theta} \sum_{j = 1}^{k} \omega_{ij}( || \mathbf{f}_i - \mathbf{f}_j ||_{2} )$, where $\theta=800$ denotes the number of Gaussian randomly sampled per iteration, and $k=5$ specifies the number of nearest neighbors for each sampled Gaussian in 3D space. $\omega_{ij} = \mathrm{exp}( - || {\mu}_{i} - {\mu}_{j} ||_{2} )$ is the distance-based weighting coefficient. 
To prevent early-stage suboptimal convergence, $\lambda$ is initialized at 0 and then increased to 1 after 4k iterations during training.

\begin{figure*}
\centering
\includegraphics[width=.97\linewidth]{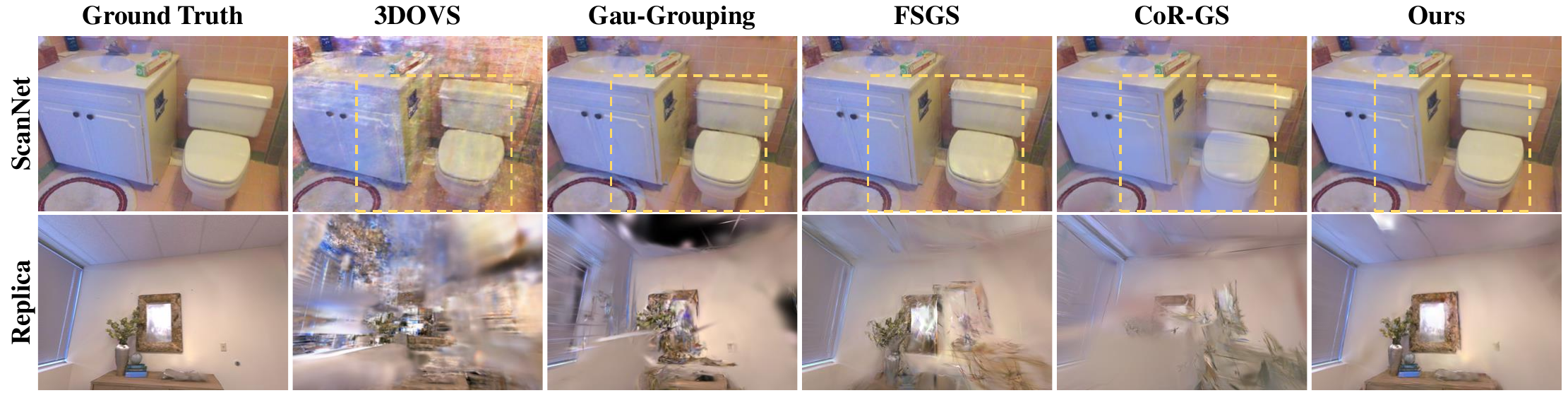}
\vspace{-3mm}
\caption{
Visual reconstruction results on novel views. Our approach achieves superior global structure and photo-realistic details. 
}
\label{fig:vis_comparison_recon}
\end{figure*}

\begin{table*}[t]
\renewcommand{\arraystretch}{1.1}
\centering 
\begin{adjustbox} {width=.97\linewidth}
\begin{tabular}{ c | p{1.3cm}<{\centering} p{1.3cm}<{\centering} p{1.3cm}<{\centering} p{1.3cm}<{\centering} p{1.3cm}<{\centering} | p{1.3cm}<{\centering} p{1.3cm}<{\centering} p{1.3cm}<{\centering} p{1.3cm}<{\centering} p{1.3cm}<{\centering} p{1.3cm}<{\centering} }
\Xhline{3\arrayrulewidth}
\multirow{2}{*}{Method} & \multicolumn{5}{c|}{Replica \cite{replica}} & \multicolumn{5}{c}{ScanNet \cite{scannet}}\\ 
& PSNR $\uparrow$ & SSIM $\uparrow$ & LPIPS $\downarrow$ & mIoU $\uparrow$ & mAcc $\uparrow$ 
& PSNR $\uparrow$ & SSIM $\uparrow$ & LPIPS $\downarrow$ & mIoU $\uparrow$ & mAcc $\uparrow$ \\ 
\hline 
3DOVS        \cite{3DOVS}       & 13.200 & 0.515 & 0.663 & 3.913  & 21.439 & 18.721 & 0.643 & 0.544 & 7.751  & 36.356 \\ 
Feature 3DGS \cite{feature3dgs} & 17.059 & 0.700 & 0.359 & 7.407 & 30.209 & 18.088 & 0.656 & 0.424 & 7.347  & 38.002 \\
LangSplat    \cite{langsplat}   & 17.715 & 0.701 & 0.420 & 3.292 & 18.621 & 18.780 & 0.663 & 0.476 & 3.339  & 24.491   \\ 
Gau-Grouping \cite{gaugrouping} & 17.297 & 0.691 & 0.376 & 15.028 & 50.422 & 19.006 & 0.675 & 0.421 & \underline{15.172} & \underline{57.355} \\
\hline 
DNGaussian   \cite{dngaussian}  & 18.513 & 0.712 & 0.402 & \underline{16.657} & \underline{54.400} & 18.392 & 0.691 & 0.484 & 13.912 & 54.814 \\
FSGS \cite{fsgs}                & 18.694 & 0.725 & 0.328 & 16.101 & 53.474 & 20.574 & 0.696 & 0.406 & 14.031 & 56.850 \\ 
CoR-GS \cite{CoRGS}             & \underline{18.955} & \underline{0.758} & \underline{0.312} & 15.933 & 52.476 & \underline{20.815} & \underline{0.721} & \underline{0.393} & 14.462 & 55.812 \\ 
\hline
Ours   
& \textbf{22.011} & \textbf{0.792} & \textbf{0.254} & \textbf{23.960} & \textbf{63.262} 
& \textbf{22.401} & \textbf{0.741} & \textbf{0.368} & \textbf{28.944} & \textbf{68.823} \\
\Xhline{3\arrayrulewidth}
\end{tabular}
\end{adjustbox}
\vspace{-3mm}
\caption{Quantitative results of reconstruction and segmentation on novel views on the Replica and ScanNet datasets, using the CLIP \cite{OpenAICLIP} for optimization with 12 training views. 
Our approach achieves {\textbf{superior performances}} across all metrics on various datasets.  
}  
\label{table:SOTA_full_CLIP}
\end{table*}

\section{Experiments}
\subsection{Experimental Setup}
\textbf{Datasets.} 
We conduct experiments on two widely-used indoor scene datasets in terms of novel view synthesis: Replica \cite{replica} and ScanNet \cite{scannet}, using a sparse-input setting with only 12 input images for training. 
Following the standard sparse-input protocol \cite{dngaussian,fsgs}, every $10$-$th$ frame from each sequence is selected for testing, while 12 evenly sampled sparse views from the remaining images are used for training. 
These 12 views collectively encompass the entire scene. 
Notably, the \textit{“inside-out”} viewing direction and limited overlap among views in sparse-input indoor scenes pose significant challenges for free-view synthesis.

\noindent 
\textbf{Evaluation Metrics.} 
We employ PSNR, SSIM \cite{ssim}, and LPIPS \cite{lpips} for evaluating image quality. 
We report segmentation performance using the mean Intersection-over-Union (mIoU) and mean pixel accuracy (mAcc).

\noindent 
\textbf{Implementation Details.} 
We implement our approach using the official PyTorch 3D Gaussian Splatting \cite{3dgs} on an NVIDIA A100 GPU. 
The dimension of the learnable semantic parameter attached to 3D Gaussians is set to 32. We modify the CUDA kernel to enable parallel differentiable rasterization of both color and semantic attributes. 
We employ CLIP ViT-B/16 \cite{OpenAICLIP} for feature extraction and SAM2 Hiera-L \cite{sam2} for region mask generation. 
More details are provided in the supplementary file.

\subsection{Comparison with State-of-the-art Methods} \label{section:comparison_SOTA}
We conduct novel view synthesis comparisons using sparse input data in indoor scenes, including two types of methods. 
\textit{i) 3D open-vocabulary segmentation methods}, 
including 3DOVS \cite{3DOVS}, Feature 3DGS \cite{feature3dgs}, LangSplat \cite{langsplat}, and Gau-Grouping \cite{gaugrouping}, are designed for 3D open world semantic understanding. 
\textit{ii) Sparse-input free-view synthesis methods} contain DNGaussian \cite{dngaussian}, FSGS \cite{fsgs}, and CoR-GS \cite{CoRGS}, which are designed to learn Gaussian radiance fields from sparse inputs, focusing primarily on appearance modeling. 
For concurrent 3D open-world segmentation, we extend these methods by adding semantic properties to Gaussian and incorporating CLIP-based semantic supervision, aligning with the semantic settings of our approach. 
We adopt consistent evaluations across all methods by employing the same sparse-input training protocol and test views.

\begin{figure*}
\centering
\includegraphics[width=.97\linewidth]{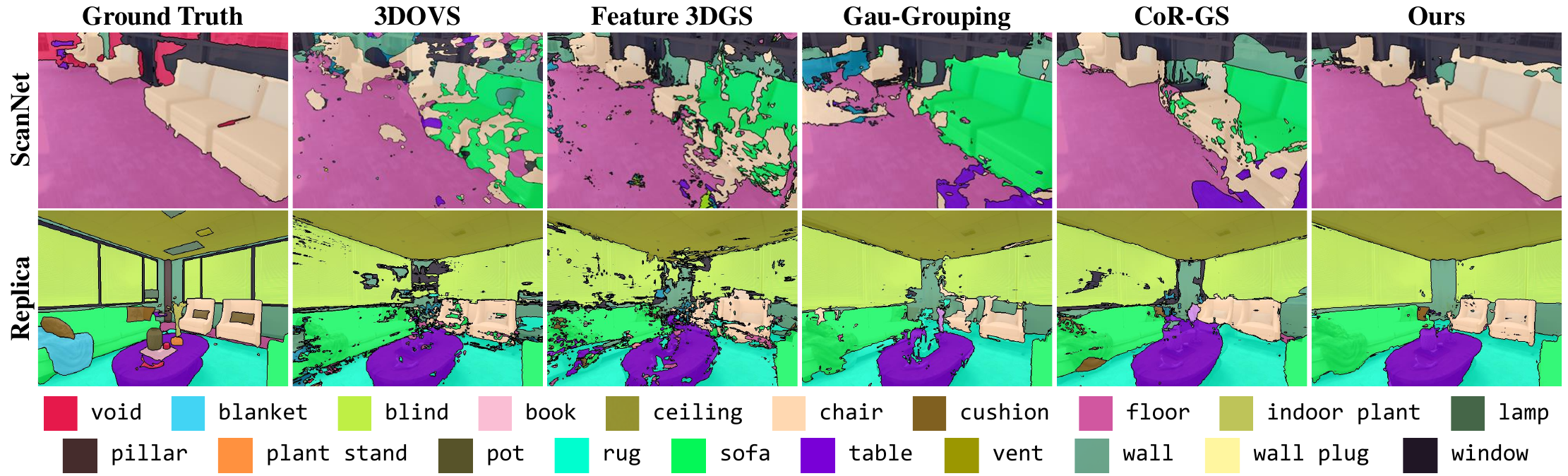}
\vspace{-3mm}
\caption{
Visual open-world segmentation results on novel views. Our approach yields more accurate and complete results. 
}
\label{fig:vis_comparison_seg}
\end{figure*}

\noindent 
\textbf{Qualitative Results.} 
\textit{i) Reconstruction.} 
As shown in Fig. \ref{fig:vis_comparison_recon}, the results from existing 3D open-vocabulary segmentation methods exhibit significant artifacts, since these methods lack sufficient view constraints. 
Although sparse-view reconstruction methods introduce geometric constraints for regularization, they struggle to recover photo-realistic details. 
In comparison, our approach incorporates scene-layout points prior and semantic consistency constraints, leading to a more detailed appearance.

\textit{ii) Open-world Segmentation.} 
In Fig. \ref{fig:vis_comparison_seg}, current methods struggle to obtain precise and complete results, since they easily encounter ambiguity arising from insufficient semantic view supervision. 
In contrast, our method delivers accurate results, benefiting from the SPC regularization that provides effective consistency constraints.

\noindent 
\textbf{Quantitative Results.} 
Table \ref{table:SOTA_full_CLIP} shows that, when optimized using the CLIP model \cite{OpenAICLIP}, our approach significantly outperforms existing methods across all metrics on both synthetic and real datasets. Competing methods struggle under sparse input conditions, due to either limited Gaussian representations or insufficient semantic consistency constraints. 
Notably, LangSplat faces considerable challenges, as it freezes Gaussian coordinates, scale, and rotation during semantic parameter optimization, which severely impedes semantic learning under sparse inputs. 
Besides, our method maintains superior results when optimized with CLIP-LSeg \cite{LSeg} (\textit{see supplementary file}), validating its generalizability.

\subsection{Ablation Studies} \label{section:ablation_study}  
\begin{figure}[!t]
\centering
\includegraphics[width=.99\linewidth]{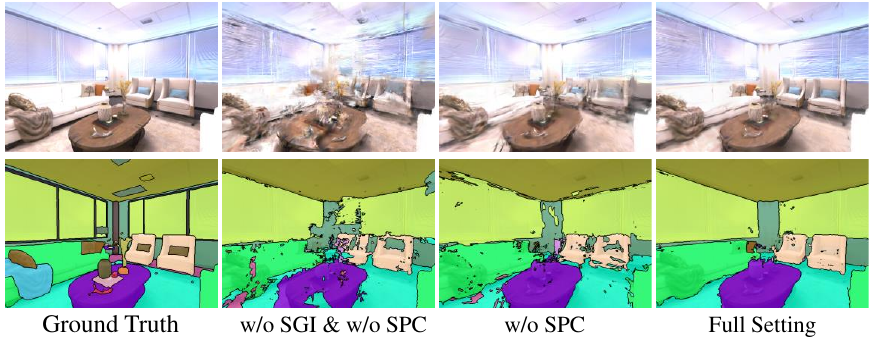}
\vspace{-3mm}
\caption{
Visual ablation comparison of key components. 
}
\label{fig:vis_ablation}
\end{figure}

\textbf{Key Components.} 
In Table \ref{table_Ablation_main}, applying the VGM-based Point Creation (PC, \#1) and Scene-layout Point Generation (SPG, \#2) notably improves reconstruction quality over setting \#0, and consequently boosts segmentation results, validating SGI's effectiveness in enhancing Gaussian representation. 
Adding the full Semantic-Prompt Consistency regularization (SPC, \#4) further yields significant segmentation gains, showing SPC’s capability to address insufficient view constraints and enhance 3D scene understanding. 
Fig. \ref{fig:vis_ablation} also shows that adding SGI can reduce artifacts, and applying SPC further enhances segmentation results.

\noindent 
\textbf{Components in SPC.} 
In Table \ref{table_Ablation_spc}, removing either the 2D view-to-view or contrastive regularization (\#b, \#c), or omitting the region boundary erosion operation in the 3D constraint (\#d), degrades performance. This validates their roles in mitigating insufficient supervision and filtering ambiguous Gaussian primitives spanning different objects.

\noindent 
\textbf{Robustness.} 
In Table \ref{table_Ablation_sparseInit}, our strategies also achieve significant gains even with sparse SfM initialization from limited training images, validating their robustness.

\begin{figure}[!t]
\centering
\includegraphics[width=.99\linewidth]{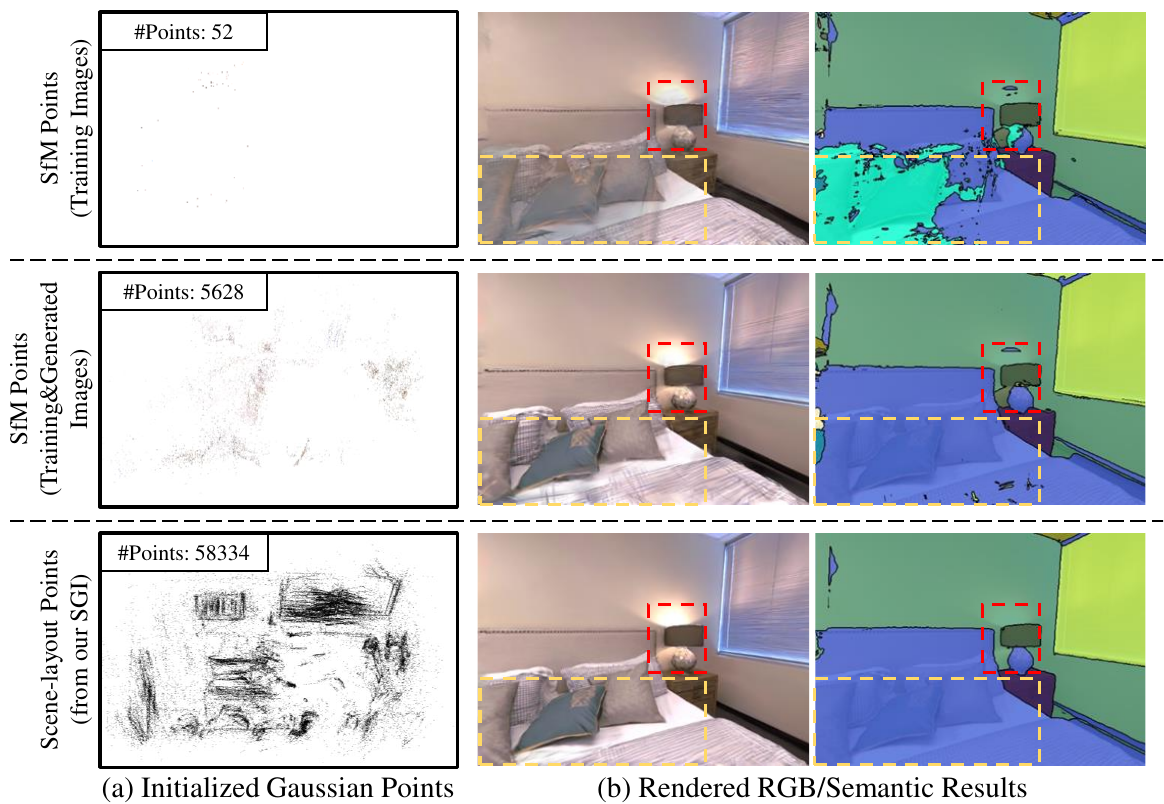}
\vspace{-3mm}
\caption{
Visual comparison of different initializations and their rendering results after optimization with these initializations. 
}
\label{fig:vis_points}
\end{figure}

\noindent 
\textbf{Initialization.} 
From top to bottom in Fig. \ref{fig:vis_points}, we can see that leveraging denser and more effective points for Gaussian initialization promotes 3DGS training that enhances scene representation, leading to improved rendered results.

\begin{table}[!t]
\renewcommand{\arraystretch}{1.1}
\label{table_Ablation_all}
\centering
\large
\begin{adjustbox} {width=\linewidth}
\begin{tabular}{ c | p{0.7cm}<{\centering} p{0.7cm}<{\centering} | p{0.7cm}<{\centering} p{0.7cm}<{\centering} | c c c c c }
\Xhline{3\arrayrulewidth}
\multirow{2}{*}{Case} & \multicolumn{2}{c|}{SGI} & \multicolumn{2}{c|}{SPC} & \multicolumn{5}{c}{Replica \cite{replica}}  \\ 
~ & PC & SPG & 2D & 3D  & PSNR  & SSIM  & LPIPS  & mIoU  & mAcc  \\
\midrule[0.8pt]  
\#0 &  &  &  &                          & 17.134 & 0.678 & 0.369 & 14.899 & 51.255 \\ 
\#1 & \checkmark  &  &  &               & 18.924 & 0.724 & 0.334 & 16.279 & 52.212 \\  
\#2 & \checkmark  & \checkmark  &  &    & 21.176 & 0.783 & 0.260 & 19.401 & 58.281 \\  
\#3 & \checkmark  & \checkmark  & \checkmark &              & 21.814 & 0.791 & 0.257 & 22.849 & 61.362 \\
\#4 & \checkmark  & \checkmark  & \checkmark & \checkmark   & \textbf{22.011} & \textbf{0.792} & \textbf{0.254} & \textbf{23.960} & \textbf{63.262} \\
\Xhline{3\arrayrulewidth}
\end{tabular}
\end{adjustbox}
\vspace{-3mm}
\caption{Ablation study on key components.}
\label{table_Ablation_main}
\end{table}

\begin{table}[!t]
\renewcommand{\arraystretch}{1.1}
\centering
\large
\begin{adjustbox} {width=\linewidth}
\begin{tabular}{ c | c | c c c c c }
\Xhline{3\arrayrulewidth}
{Case} & {Configuration} & PSNR  & SSIM  & LPIPS  & mIoU  & mAcc  \\
\midrule[0.8pt] 
\#a & Ours (Full setting)  & 22.011 & 0.792 & 0.254 & 23.960 & 63.262 \\
\hline
\#b & w/o View-to-View Regular.  & 21.683 & 0.789 & 0.257 & 21.244 & 60.914 \\
\#c & w/o Contrastive Regular.   & 21.879 & 0.792 & 0.256 & 22.322 & 60.086 \\
\#d & w/o Boundary Erosion & 21.950 & 0.792 & 0.254 & 22.964 & 61.706 \\ 
\Xhline{3\arrayrulewidth}
\end{tabular}
\end{adjustbox}
\vspace{-3mm}
\caption{Ablation results of our SPC with different settings.} 
\label{table_Ablation_spc} 
\end{table}

\begin{table}[!t]
\renewcommand{\arraystretch}{1.1}
\centering
\large
\begin{adjustbox} {width=\linewidth}
\begin{tabular}{ c | c | c c c c c }
\Xhline{3\arrayrulewidth}
Case & {Configuration}         & PSNR  & SSIM  & LPIPS  & mIoU  & mAcc  \\
\midrule[0.8pt]  
\#a  & SfM Points Initialization & 17.134 & 0.678 & 0.369 & 14.899 & 51.255 \\   
\#b  & \#a + SGI     & 20.680 & 0.774 & 0.270 & 18.169 & 56.928 \\   
\#c  & \#b + SPC     & 21.589 & 0.786 & 0.260 & 20.766 & 61.000 \\   
\Xhline{3\arrayrulewidth} 
\end{tabular}
\end{adjustbox}
\vspace{-3mm}
\caption{Results of our strategies using SfM point initialization.} 
\label{table_Ablation_sparseInit}  
\end{table}

\section{Conclusion}
This paper presents SPC-GS, a novel approach to address the challenges of GS-based indoor open-world free-view synthesis with sparse views. 
In SPC-GS, the proposed Scene-layout-based Gaussian Initialization (SGI) effectively enhances Gaussian representation by providing an instructive point initialization, consequently promoting semantic learning. 
Additionally, the Semantic-Prompt Consistency (SPC) regularization mitigates limited supervision by introducing additional view-consistent semantic constraints, improving coherent segmentation performance. 
Experimental results show that our approach significantly outperforms SOTA methods in reconstruction and 3D open-world segmentation tasks across synthetic and real-world datasets with various CLIP models.

\clearpage
{
    \small
    \bibliographystyle{ieeenat_fullname}
    \bibliography{main}
}

\clearpage
\setcounter{page}{1}
\maketitlesupplementary
\appendix

In this appendix, we present additional materials to support and extend the findings and observations presented in the main body of this paper. 

\begin{itemize}
    \item Section \ref{SGI_detals} presents more details of the Scene-layout-based Gaussian Initialization (SGI) strategy. 
    \item Section \ref{Implementation_detals} elaborates on additional experimental details of our approach. 
    \item Section \ref{More_Experiements} offers extended experimental analyses to showcase the effectiveness of our method. 
    \item Section \ref{More_Visualization_results} presents more qualitative results to facilitate better visual comparisons. 
    \item Section \ref{Scene_Qualitative_results} provides per-scene quantitative evaluation results for more comprehensive comparisons.  
\end{itemize}

\section{More Details of Scene-layout-based Gaussian Initialization (SGI)} \label{SGI_detals}
Our proposed SGI strategy includes two main components: \textit{VGM-based Point Creation} and \textit{Scene-layout Point Generation}. 
\textit{i)} The first component aims to produce denser SfM points by leveraging additional view-changed images that are generated from the original sparse training views. 
\textit{ii)} The second component yields an instructive scene-layout point distribution for enhanced Gaussian initialization.

\begin{figure}[!t]
\centering
\includegraphics[width=\linewidth]{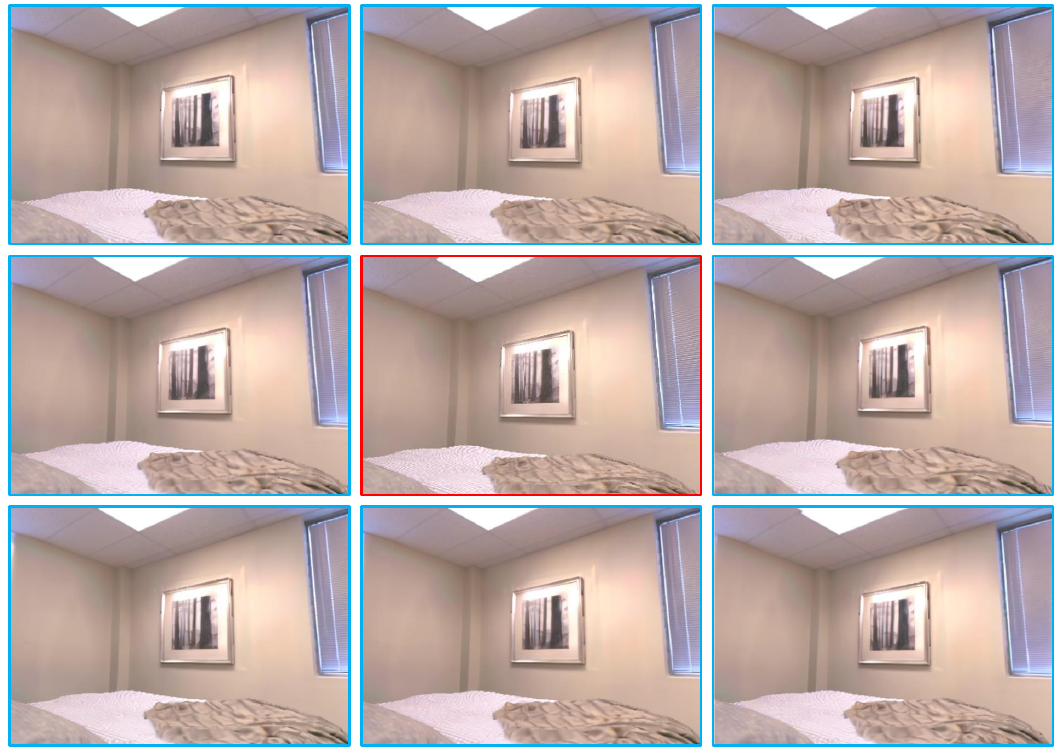}
\caption{
Illustration of training image (red border in the center) and corresponding generated images (blue border). 
}
\label{fig:vis_generatedViews}
\end{figure}

\begin{figure*}
\centering
\includegraphics[width=\linewidth]{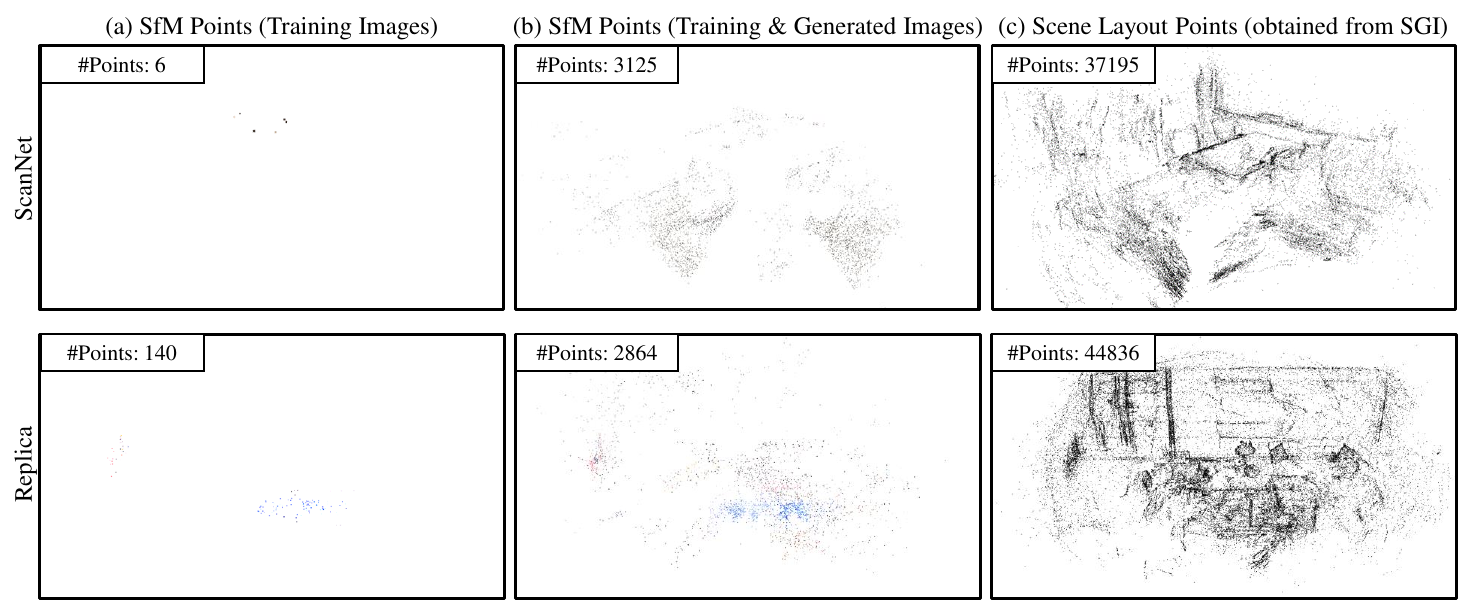}
\caption{
Visualization of Gaussian point distributions of different scenes: 
(a) Sparsely distributed SfM points from limited training views; 
(b) Denser SfM points distribution using generated images and training images; 
(c) Instructive point distribution obtained via our Scene-layout-based Gaussian Initialization (SGI) strategy, demonstrating improved scene layout coverage that enhances reconstruction quality and consequently improves segmentation accuracy. 
}
\label{fig:vis_Points}
\end{figure*}

Specifically, in the VGM-based Point Creation process, eight neighboring views \(\{\tilde{I}_i^j\}_{j=1}^{8}\) are generated for each original training image \(I_i\) using the image-to-video generation mode of the advanced video generation model MotionCtrl \cite{motionctrl}. 
Fig. \ref{fig:vis_generatedViews} shows these additional view-changed images (denoted by the blue border) alongside the original training view (indicated by the red border). 
Subsequently, the generated images are combined with the original training images for SfM processing, yielding denser initialized SfM points. 
As shown in Table \ref{table:number_gaussians}, we can see that sparse-input training images alone yield limited SfM points. 
Augmenting the training images with view-changed images increases the amount of SfM points. 
Furthermore, the Scene-layout Point Generation further produces a scene-layout-wise Gaussian point distribution, serving as an enhanced and instructive initialization prior. 
Moreover, these results also can be found in Fig. \ref{fig:vis_Points}.

In general, these results demonstrate that our full SGI framework effectively provides dense and instructive points for Gaussian initialization, which promotes scene Gaussian representation, and consequently, enhances semantic Gaussian learning in sparse-input scenarios.

\begin{table}[!t] 
\renewcommand{\arraystretch}{1.1}
\centering 
\begin{adjustbox} {width=\linewidth}
\begin{tabular}{ c | c | c | c }
\Xhline{3\arrayrulewidth}
\multirow{2}{*}{Scene} & \multirow{2}{*}{SfM Points}  & SfM Points with & Scene Layout \\
~ & ~ & Generated Views & Points from SGI \\ 
\hline
Room0     & 122       & 6527     & 57524     \\
Room1     & 52        & 5628     & 58334     \\
Room2     & 22        & 2659     & 54308     \\
Office0   & 474       & 5452     & 41989     \\
Office2   & 140       & 2864     & 44836     \\
Office4   & 110       & 3621     & 29115     \\
Scene0004 & 70        & 4027     & 64016     \\
Scene0389 & 19        & 1207     & 22901     \\
Scene0494 & 6         & 3125     & 37195     \\
Scene0693 & 48        & 2911     & 44060     \\                
\Xhline{3\arrayrulewidth}
\end{tabular}
\end{adjustbox}
\caption{
Number of Gaussian points in various settings. 
``SfM Points" refers to the initialized Gaussian points derived from the Structure-from-Motion (SfM) algorithm using sparse training images. 
``SfM Points with Generated Views" indicates the initialized SfM Gaussian points derived from the SfM algorithm using training and generated images. 
``Scene Layout Points from SGI" denotes the Gaussian points obtained from the Scene-layout Gaussian Initialization (SGI) strategy, which are treated as initialized Gaussian points to optimize the Gaussian radiance field. 
} 
\label{table:number_gaussians}
\end{table}

\section{Elaborated Experimental Details} \label{Implementation_detals}

\subsection{Datasets} 
To evaluate the performance of our approach, we conduct sparse-input open-world free-view synthesis experiments on two widely-used benchmark indoor scene datasets: Replica \cite{replica} and ScanNet \cite{scannet}.

Replica is a synthetic scene dataset comprising diverse, high-quality indoor room-scale environments. Each scene features photo-realistic textures, dense geometry, and semantic classes. 
For experiment evaluations, six commonly evaluated scenes from Replica are utilized: room0, room1, room2, office0, office2, and office4. 
Following 3DOVS \cite{3DOVS}, 45 categories are used for text queries: candle, book, vent, box, comforter, switch, bin, plant stand, bed, desk organizer, rug, bench, vase, bottle, ceiling, blanket, bowl, camera, wall, blinds, pillar, sculpture, tablet, chair, lamp, indoor plant, cabinet, stool, table, cushion, panel, plate, basket, pot, tissue paper, nightstand, sofa, window, picture, wall plug, tv screen, shelf, door, floor, clock. 

ScanNet is a real-world indoor scene dataset that includes semantic segmentation labels and camera poses provided by BundleFusion \cite{Bundlefusion}. For evaluation, four scenes are selected from ScanNet: scene0004$\underline{~}$00, scene0389$\underline{~}$00, scene0494$\underline{~}$00, and scene0693$\underline{~}$00. 
The commonly used 20 categories defined by ScanNet are used for text queries: wall, floor, cabinet, bed, chair, sofa, table, door, window, bookshelf, picture, counter, desk, curtain, refrigerator, shower curtain, toilet, sink, bathtub, other furniture. 

Following the sparse-input experimental protocol outlined in \cite{dngaussian,fsgs}, 
we select every 10-$th$ image in the sequence from each scene as the testing view, resulting in 22 to 30 testing images per scene for evaluation.   
From the remaining images, we uniformly sample 12 views to construct the sparse training set. 
The resolution of all images is set to 640 $\times$ 448. 
During training, RGB images from the training set are utilized for scene reconstruction, with CLIP-derived semantic features applied for semantic Gaussian learning. For evaluation, only the RGB images and ground-truth semantic labels from the testing set are utilized.

\begin{table*}[t]
\renewcommand{\arraystretch}{1.1}
\centering 
\begin{adjustbox} {width=\linewidth}
\begin{tabular}{ c | p{1.4cm}<{\centering} p{1.4cm}<{\centering} p{1.4cm}<{\centering} p{1.4cm}<{\centering} p{1.4cm}<{\centering} | p{1.4cm}<{\centering} p{1.4cm}<{\centering} p{1.4cm}<{\centering} p{1.4cm}<{\centering} p{1.4cm}<{\centering} p{1.4cm}<{\centering} }
\Xhline{3\arrayrulewidth}
\multirow{2}{*}{Method} & \multicolumn{5}{c|}{Replica \cite{replica}} & \multicolumn{5}{c}{ScanNet \cite{scannet}}\\ 
& PSNR $\uparrow$ & SSIM $\uparrow$ & LPIPS $\downarrow$ & mIoU $\uparrow$ & mAcc $\uparrow$ 
& PSNR $\uparrow$ & SSIM $\uparrow$ & LPIPS $\downarrow$ & mIoU $\uparrow$ & mAcc $\uparrow$ \\ 
\hline
3DOVS        \cite{3DOVS}       & 13.326 & 0.520 & 0.656 & 8.859  & 41.909 & 18.836 & 0.652 & 0.535 & 24.001 & 72.863  \\
Feature 3DGS \cite{feature3dgs} & 18.154 & 0.708 & 0.339 & 16.357 & 64.955 & 19.702 & 0.678 & 0.403 & 21.053 & 72.722  \\ 
Gau-Grouping \cite{gaugrouping} & 17.787 & 0.709 & 0.350 & 20.252 & 66.519 & 19.189 & 0.682 & 0.419 & 33.549 & 73.624 \\ 
\hline 
DNGaussian   \cite{dngaussian}  & 19.964 & 0.749 & 0.370 & 23.692 & 70.705 & 20.231 & 0.706 & 0.451 & 32.122 & 74.895  \\  
FSGS \cite{fsgs}                & 20.371 & 0.768 & 0.285 & 23.721 & 71.235 & 21.875 & 0.730 & 0.386 & 34.610 & 77.840  \\ 
CoR-GS \cite{CoRGS}             & 20.066 & 0.779 & 0.290 & 22.496 & 68.901 & 21.801 & 0.735 & 0.388 & 34.170 & 77.944  \\ 
\hline
Ours  
& \textbf{22.123} & \textbf{0.800} & \textbf{0.248} & \textbf{29.173} & \textbf{75.482} 
& \textbf{23.042}  & \textbf{0.755}  & \textbf{0.359} & \textbf{50.271} & \textbf{83.584}  \\  
\Xhline{3\arrayrulewidth}
\end{tabular}
\end{adjustbox}
\caption{Quantitative comparison of reconstruction and segmentation results on novel views in Replica and ScanNet datasets, using the CLIP-LSeg \cite{LSeg} to optimize Gaussian semantic attributes with 12 training views. 
Our approach achieves {\textbf{superior results}} across all metrics on various datasets. 
} 
\label{table:SOTA_full_LSeg}
\end{table*}

\subsection{Implementation Details} 
\textbf{Data Prerocessing.}
Before training, CLIP features are pre-computed offline, following prior methods \cite{3DOVS,feature3dgs}. 
The generated images of each training view are obtained using MotionCtrl \cite{motionctrl} under the image-to-video mode.

\noindent
\textbf{Training.} 
We set the learning rate as 0.0025 for Gaussian semantic parameters, while convolution layers $\omega_{f}$ and $\omega_{s}$ are optimized using Adam with the learning rate of 0.0005. 
The Outlier Gaussian Primitive Removal (OGR) strategy is implemented every 3k iterations. 
We first train our model to derive a scene-layout Gaussian point distribution through 10k iterations. These points are then used to initialize Gaussian positions, color attributes (using zero-order spherical harmonics), and semantic attributes, which undergo training for 10k iterations. 
The entire process costs 45 minutes on average per scene on one A100 GPU.

\noindent
\textbf{Inference.} 
During inference, we project 3D Gaussians onto the 2D plane, concurrently producing rendered RGB images and rendered semantic features in novel views. 
Following previous 3D open-world segmentation methods \cite{3DOVS, langsplat}, a set of text queries are utilized to calculate the cosine similarity between these text features and the rendered features, generating the open-vocabulary segmentation results. 
Our approach achieves a rendering speed of over 300 FPS at a resolution of 640 × 448.

\section{Additional Analyses} \label{More_Experiements}
In this section, we present additional experimental results to further validate the robustness of our proposed method. Specifically, we conduct comparative analyses using various vision-language foundation models, such as CLIP-LSeg \cite{LSeg} and APE \cite{ape}, which have been adopted in prior works \cite{feature3dgs,goi} for optimizing Gaussian semantic attributes. 
Furthermore, we include a comparative study with ViewCrafter \cite{viewcrafter}. 
Additionally, we provide more comprehensive ablation studies to evaluate the effectiveness of different components within our framework.

\subsection{Results using CLIP-LSeg Model}
To assess the effectiveness and generalizability of our approach, we utilize CLIP-LSeg \cite{LSeg} for semantic Gaussian optimization and report the quantitative results in Table \ref{table:SOTA_full_LSeg}.

\textit{i) Comparison with 3D Open-vocabulary Segmentation Methods. }  
As shown in Table \ref{table:SOTA_full_LSeg}, we see that our method consistently surpasses competitors across all metrics on various datasets when using CLIP-LSeg \cite{LSeg} to optimize Gaussian semantic attributes, while other approaches encounter significant challenges under the sparse input condition. 

Specifically, the state-of-the-art method Gau-Grouping \cite{gaugrouping} demonstrates relatively low reconstruction quality and limited performance in open-world segmentation. 
This is because it uses sparse SfM points for Gaussian initialization, which hampers its ability to represent complex indoor scenes, resulting in inferior reconstruction quality and impaired segmentation precision. 
Additionally, Gau-Grouping only applies supervision to Gaussians within sparse training views, leading to the under-optimization problem. As a result, this method tends to overfit the training views while producing less accurate results for novel viewpoints. 
In comparison, our method utilizes dense scene-layout points and semantic-prompt consistency constraints, yielding improvements of 3.97 PSNR and 8.92\% mIoU over the second-best method on the Replica benchmark. 
These results highlight the effectiveness of our approach for open-world free-view synthesis with sparse inputs.

\textit{ii) Comparison with Sparse-input Free-view Synthesis Methods.}  
In Table \ref{table:SOTA_full_LSeg}, although methods like DNGaussian, FSGS, and CoR-GS achieve improved novel view quality by incorporating additional depth or color regularizations to optimize sparse-view Gaussian radiance fields, our approach consistently surpasses them in both reconstruction quality and semantic understanding accuracy. 
These improvements can be attributed to the effectiveness of our SGI strategy in enhancing Gaussian representation and the SPC regularization in boosting segmentation accuracy. 

In general, these quantitative results validate the effectiveness and generalizability of our approach in 3D indoor open-world free-view synthesis from sparse input images, optimized across different CLIP models. 
Moreover, the qualitative results using CLIP-LSeg to optimize Gaussian semantic attributes. can be found in Section \ref{More_Visualization_results} and Fig. \ref{fig:vis_comparison_lseg_supple}.

\subsection{Results using APE Model}
Following GOI \cite{goi}, we utilize the same Aligning and Prompting Everything All at Once (APE) model \cite{ape} to extract 2D semantic features from training views, which are treated as the 2D ground truth semantic features for optimizing Gaussian semantic attributes. 
Since GOI only supports single-query segmentation results using a vocabulary, rather than generating a relevancy map at a time as outlined in its paper \cite{goi}, we adopt this single-query segmentation setting to produce sparse-input open-world free-view synthesis results. 
In line with GOI's implementations, we employ the sparse-input training images for optimization, and then evaluate the single-query performance in novel views for comparisons.

\begin{figure}
\centering
\includegraphics[width=\linewidth]{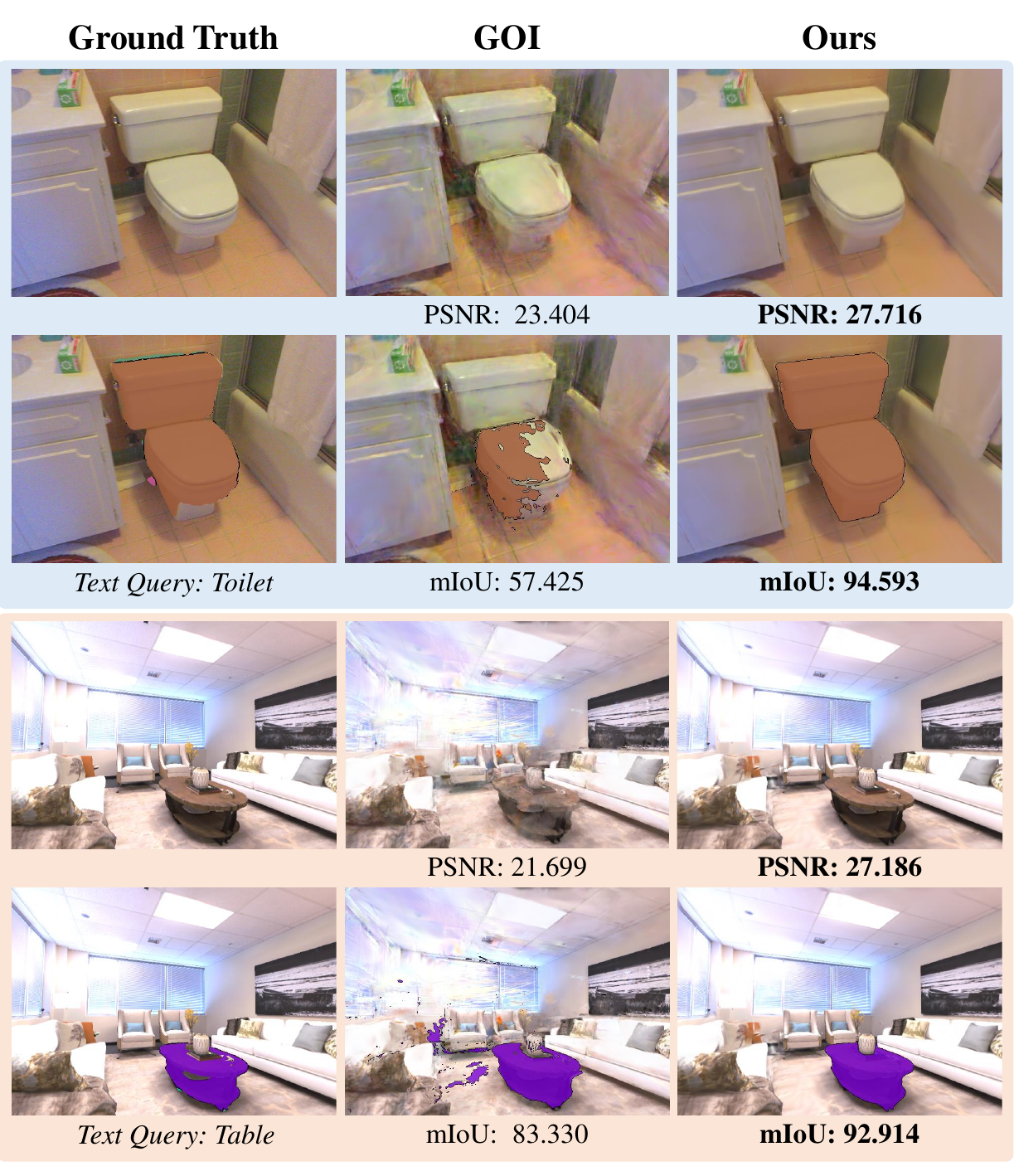}
\caption{
Comparison of reconstruction (rows 1, 3) and open-vocabulary segmentation (rows 2, 4) results on novel views across diverse scenes from the ScanNet and Replica datasets using APE \cite{ape} for optimizing Gaussian semantic attributes.  
Compared to GOI, our approach demonstrates photo-realistic appearance details and more complete segmentation results. 
}
\label{fig:vis_comparison_ape}
\end{figure}

The single-query segmentation results on ScanNet and Replica test data are illustrated in Fig. \ref{fig:vis_comparison_ape}. 
It can be seen that GOI generates vague reconstruction results in novel viewpoints. 
This can be attributed to its limited Gaussian representation stemming from sparse Gaussian point initialization. 
Consequently, inheriting the bottleneck of inferior Gaussian representation, GOI easily faces challenges and produces incomplete and noisy segmentation results under sparse input conditions.

\subsection{Comparison with ViewCrafter} 
\begin{figure}[!t]
\centering
  \includegraphics[width=\linewidth]{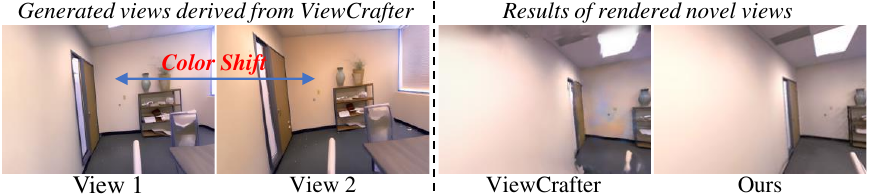}
\caption{Visual results. Compared to ViewCrafter, our approach exhibits more photo-realistic appearance details. } 
\label{fig_viewcrafter}
\end{figure}

ViewCrafter \cite{viewcrafter} employs a pre-trained video generation model to synthesize additional views, aiming to enhance sparse-view 3DGS optimization. 
However, it inherits the \textit{color shift} artifacts commonly associated with diffusion models, often generating reasonable structures with inaccurate color representations, as shown in the left part of Fig. \ref{fig_viewcrafter}.
As a result, its reconstruction performance is inferior, as shown in Table \ref{table_viewcrafter} and Fig. \ref{fig_viewcrafter}.

\begin{table}[t]
\renewcommand{\arraystretch}{1.1}
\centering 
\begin{adjustbox} {width=\linewidth}
\begin{tabular}{ c | c c c | c c c }
\Xhline{3\arrayrulewidth}
\multirow{2}{*}{Method} & \multicolumn{3}{c|}{Replica \cite{replica}}  & \multicolumn{3}{c}{ScanNet \cite{scannet}} \\ 
~  & PSNR  & SSIM  & LPIPS  & PSNR  & SSIM  & LPIPS  \\  
\hline 
ViewCrafter \cite{viewcrafter} & 19.207 & 0.762 & 0.325 & 17.262 & 0.697 & 0.468  \\
Ours        & \textbf{22.011} & \textbf{0.792} & \textbf{0.254} & \textbf{22.401} & \textbf{0.741} & \textbf{0.368} \\ 
\Xhline{3\arrayrulewidth}
\end{tabular}
\end{adjustbox}
\caption{
Quantitative results of reconstruction on novel views. 
} 
\label{table_viewcrafter}
\end{table}

\subsection{More Ablation Studies} 
In this section, we present more ablation analyses of our approaches for a more comprehensive analysis.

\begin{table}[!t]
\renewcommand{\arraystretch}{1.1}
\centering
\large
\begin{adjustbox} {width=\linewidth}
\begin{tabular}{ c | c | c c c c c }
\Xhline{3\arrayrulewidth}
\multirow{2}{*}{Case} & \multirow{2}{*}{Configuration} & \multicolumn{5}{c}{Replica \cite{replica}}  \\ 
~  &  ~  & PSNR $\uparrow$ & SSIM $\uparrow$ & LPIPS $\downarrow$ & mIoU $\uparrow$ & mAcc $\uparrow$ \\ 
\midrule[0.8pt] 
\#a & Ours (Full Setting)     & \textbf{22.011} & \textbf{0.792} & \textbf{0.254} & \textbf{23.960} & \textbf{63.262} \\
\midrule[0.5pt] 
\#b & w/o OGR                 & 21.889 & 0.790 & 0.258 & 22.501 & 62.085 \\ 
\#c & w $\mathcal{L}_{GR}$    & 20.296 & 0.761 & 0.304 & 19.864 & 57.812 \\
\#d & w 2D CLIP for $\mathcal{L}_{inter}$    & 21.906 & 0.790 & 0.259 & 23.139 & 62.596 \\ 
\Xhline{3\arrayrulewidth}
\end{tabular}
\end{adjustbox}
\caption{Further ablation results of our approach with various settings.} 
\label{table_Ablation_supple} 
\end{table}

\begin{figure}[!t]
\centering
\includegraphics[width=\linewidth]{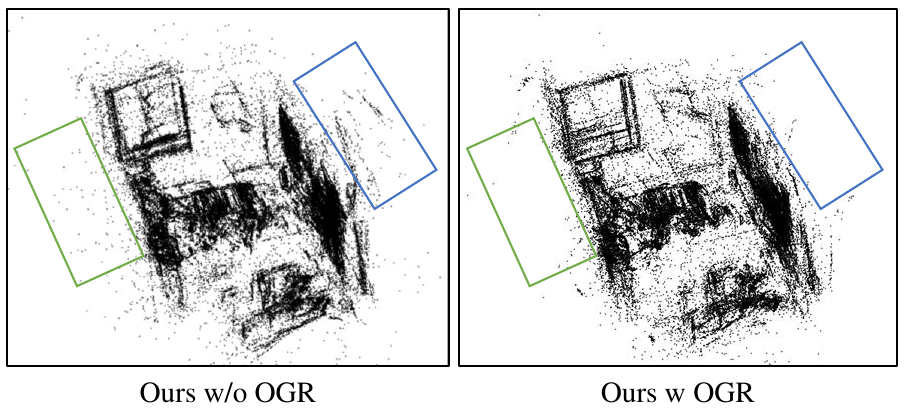}
\caption{
Comparison of Gaussian point clouds with and without Outlier Gaussian Primitive Removal (OGR) strategy. 
Applying the OGR can reduce outlier Gaussian primitives, facilitating the optimization of Gaussian representations and enhancing results. 
}
\label{fig:vis_OGR}
\end{figure}

\noindent
\textbf{Effectiveness of OGR.} 
To assess the effectiveness of the Outlier Gaussian Primitive Removal (OGR), we present ablation results in Fig. \ref{fig:vis_OGR} and \#b of Table \ref{table_Ablation_supple}. 
As depicted in Fig. \ref{fig:vis_OGR}, the omission of OGR results in an increased presence of outlier Gaussian primitives. 
These excessive outlier primitives, when used for subsequent Gaussian initialization, complicate the optimization stage, leading to performance degradation as shown in configuration \#b of Table \ref{table_Ablation_supple}. 
These findings demonstrate the efficacy of the OGR strategy in mitigating outlier proliferation, preserving Gaussian representation quality, and enhancing rendering results.

\begin{figure}[!t]
\centering
\includegraphics[width=\linewidth]{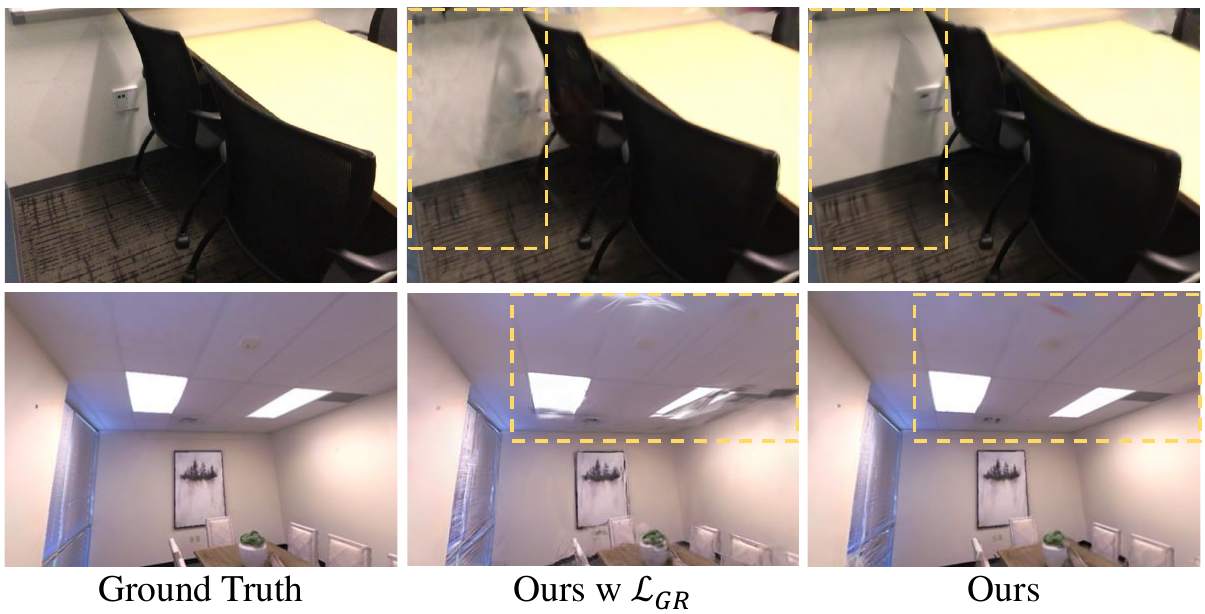}
\caption{
Visual ablation results for using generated-view color supervision $\mathcal{L}_{GR}$. Incorporating color supervision $\mathcal{L}_{GR}$ (``Ours w $\mathcal{L}_{GR}$") provided from generated RGB images produces inferior results compared to without using $\mathcal{L}_{GR}$ (``Ours"). 
}
\label{fig:vis_generated_color}
\end{figure}

\noindent
\textbf{Impact of Generated-view Color Supervision.}
To investigate the impact of color supervision $\mathcal{L}_{GR}$ provided by generated images, we present the ablation results in \#c of Table \ref{table_Ablation_supple} and Fig. \ref{fig:vis_generated_color}. 
It can be observed that the addition of color supervision $\mathcal{L}_{GR}$ leads to performance degradation. 
This occurs because the reconstruction task is highly sensitive to erroneous color pixels in generated images, as these inaccuracies can significantly degrade reconstruction quality, with even minor color inconsistencies affecting reconstruction quality, as illustrated in Fig. \ref{fig:vis_generated_color}.

\noindent
\textbf{Analysis of the Pseudo Supervision Signals.}
To analyze the effectiveness of using self-rendered semantic representations from training views for supervising pseudo views in $\mathcal{L}_{inter}$ (\ie Eq. \eqref{eq_interloss}), we replace them with the 2D CLIP-derived semantics from training views. 
As shown in \#d of Table \ref{table_Ablation_supple}, this replacement leads to degraded performance. This demonstrates the advantage of self-rendered semantic representations derived from the 3D radiance fields exhibit superior coherent and reliable information over the 2D representations, enhancing overall optimization.

\noindent
\textbf{Analysis of the View Constraint in SGI.}
We examine the effect of view constraints on 3D Gaussian densification within the SGI by utilizing only two components: the vanilla reconstruction loss $\mathcal{L}_C$ and semantic loss $\mathcal{L}_S$ (Eq. \eqref{semantic_loss}), denoted as SGI* in Table \ref{table_Ablation_SGI}.  
The experimental results demonstrate that our SGI strategy also achieves significant improvements, while the additional application of SPC further enhances performance. 
These findings indicate that our SGI can generate effective scene-layout Gaussian distributions without relying on specific view constraints.

\begin{table}[!t]
\renewcommand{\arraystretch}{1.1}
\centering
\large
\begin{adjustbox} {width=\linewidth}
\begin{tabular}{ c | c | c c c c c }
\Xhline{3\arrayrulewidth}
{Case} & {Configuration}     & PSNR $\uparrow$ & SSIM $\uparrow$ & LPIPS $\downarrow$ & mIoU $\uparrow$ & mAcc $\uparrow$ \\ 
\midrule[0.8pt] 
\#a & Baseline               & 17.134 & 0.678 & 0.369 & 14.899 & 51.255 \\
\#b & \#a + SGI*             & 20.914 & 0.782 & 0.284 & 18.865 & 57.032 \\ 
\#c & \#b + SPC              & 21.806 & 0.792 & 0.254 & 23.202 & 62.057 \\  
\Xhline{3\arrayrulewidth}
\end{tabular}
\end{adjustbox}
\caption{Ablation results of our SGI strategy incorporating only vanilla reconstruction loss $\mathcal{L}_C$ and semantic loss $\mathcal{L}_S$ (Eq. \eqref{semantic_loss}).}
\label{table_Ablation_SGI} 
\end{table}

\noindent 
\textbf{Analysis of the Hyperparameter for Region Boundary Erosion.}
Table \ref{table_Ablation_Erosion} presents an experimental analysis of the erosion hyperparameter used for Region Boundary Erosion.  
Specifically, the configuration $3 \times 3$ indicates that a pixel $\mathbf{M}'(x, y)$ is retained as ``True" only if all pixels within its $3 \times 3$ neighborhood in the input original mask $\mathbf{M}$ are ``True"; otherwise, it is assigned ``False". This operation effectively contracts region boundaries inward. 

The experimental results demonstrate that increasing the size of the erosion hyperparameter, \ie using a larger kernel, helps reduce the number of ambiguous Gaussian primitives near object boundaries, thereby improving performance. 
However, excessive erosion diminishes the number of available Gaussians for optimization, potentially limiting performance. 
Based on these empirical findings, we adopt the $5 \times 5$ configuration as the final parameter choice.

\noindent 
\textbf{Analysis of the Number of Point Prompts in ISP.}
In Table \ref{table_Ablation_ISP}, we evaluate the performance across varying quantities of point prompts used to establish region mask correspondences. 
The experimental results demonstrate that increasing point prompt quantities yields performance comparable to the baseline configuration. This finding suggests that the \textit{iterative stochastic design} in the base ISP effectively generates a uniform distribution of points across the entire image space as the training processes. 
This enables efficient construction of region mask correspondences across diverse image regions. 
Overall, given the additional training time required for incorporating more point prompts and their marginal performance gains, we opt not to include further point sampling in the final framework.

\begin{table}[!t]
\renewcommand{\arraystretch}{1.1}
\centering
\large
\begin{adjustbox} {width=.98\linewidth}
\begin{tabular}{ c | c | c c c c c }
\Xhline{3\arrayrulewidth}
{Case} & {Configuration} & PSNR $\uparrow$ & SSIM $\uparrow$ & LPIPS $\downarrow$ & mIoU $\uparrow$ & mAcc $\uparrow$ \\ 
\midrule[0.8pt] 
\#a & 1 $\times$ 1            & 21.876 & 0.792 & 0.255 & 23.270 & 61.409 \\ 
\#b & 3 $\times$ 3            & 21.901 & 0.791 & 0.255 & 23.474 & 62.485 \\ 
\#c & 5 $\times$ 5            & \textbf{22.011} & \textbf{0.792} & \textbf{0.254} & \textbf{23.960} & \textbf{63.262} \\
\#d & 7 $\times$ 7            & 21.920 & 0.792 & 0.255 & 23.388 & 62.314 \\ 
\#e & 9 $\times$ 9            & 21.950 & 0.792 & 0.255 & 23.182 & 62.349 \\ 
\Xhline{3\arrayrulewidth}
\end{tabular}
\end{adjustbox}
\caption{Analysis of the erosion hyperparameter for Region Boundary Erosion.} 
\label{table_Ablation_Erosion} 
\end{table}

\begin{table}[!t]
\renewcommand{\arraystretch}{1.1}
\centering
\large
\begin{adjustbox} {width=\linewidth}
\begin{tabular}{ c | c | c c c c c }
\Xhline{3\arrayrulewidth}
Point Prompts & \#Train & PSNR $\uparrow$ & SSIM $\uparrow$ & LPIPS $\downarrow$ & mIoU $\uparrow$ & mAcc $\uparrow$ \\ 
\midrule[0.8pt]  
Number=1 (Base)  & $\sim$45mins   & 22.011 & 0.792 & 0.254 & 23.960 & 63.262 \\
Number=2         & $\sim$50mins   & 22.044 & 0.792 & 0.254 & 24.089 & 63.365 \\ 
Number=3         & $\sim$55mins   & 22.034 & 0.792 & 0.254 & 24.364 & 63.440 \\ 
\Xhline{3\arrayrulewidth}
\end{tabular}
\end{adjustbox}
\caption{Performance for the Iterative Stochastic Prompting (ISP) approach with different settings.} 
\label{table_Ablation_ISP} 
\end{table}

\noindent 
\textbf{Analysis of Lower-Order Spherical Harmonics (SH).}
As shown in Table \ref{table_sh}, employing lower-order SH under generated view constraints (with $\mathcal{L}_{GR}$) partially alleviates the performance drop in evaluation metrics. However, a noticeable degradation still occurs, primarily due to color inaccuracies in the generated images.

\begin{figure}[!t]
\centering
  \includegraphics[width=\linewidth]{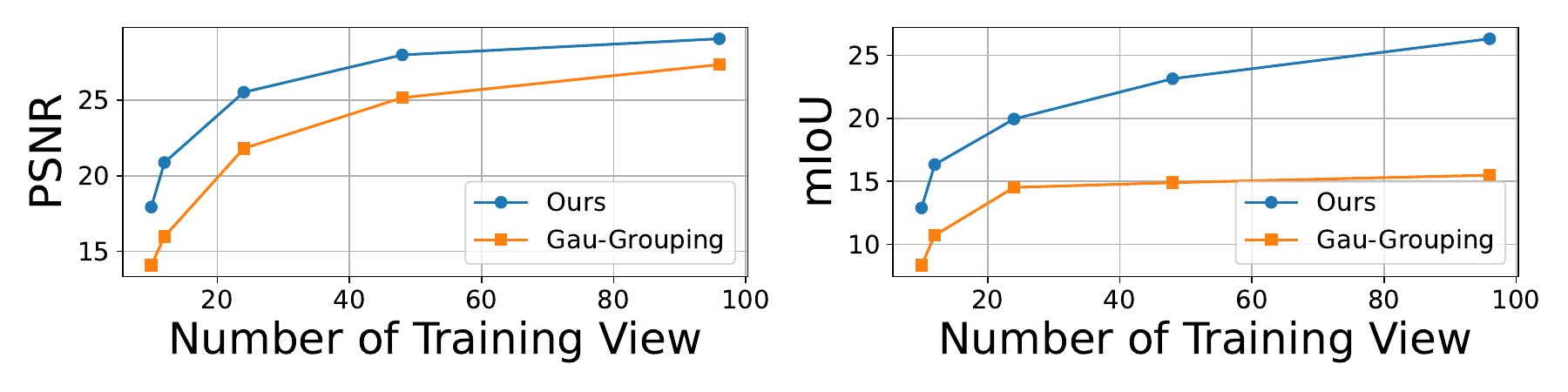}
\caption{Results using 10, 12, 24, 48, and 96 training views.}
\label{fig_TrainViews}
\end{figure}

\noindent 
\textbf{Analysis of Varying Numbers of Training Views.}
As illustrated in Fig. \ref{fig_TrainViews}, our method demonstrates strong robustness when using varying numbers of training views in Room0, ranging from very sparse (fewer than 10 views, where COLMAP fails) to dense view configurations. 
The superior segmentation performance is attributed to the proposed semantic-prompt consistency regularization strategy.

\subsection{Limitation Analysis and Future Work}  
While our SPC-GS framework demonstrates notable advantages in sparse-input open-world free-view synthesis, it is currently limited to static 3D scenes, as it does not incorporate dynamic Gaussian modeling or time-dependent optimization mechanisms.  
Future research directions could explore the extension of our method to dynamic open-world free-view synthesis, enabling temporal scene modeling.

\begin{table}[!t]
\renewcommand{\arraystretch}{1.1}
\centering 
\begin{adjustbox} {width=0.98\linewidth}
\begin{tabular}{ c | c | c c c c c } 
\Xhline{3\arrayrulewidth}
\multirow{2}{*}{Loss} & \multirow{2}{*}{SH Setting} & \multicolumn{5}{c}{Room0 Scene}  \\ 
~  &  ~  & PSNR  & SSIM  & LPIPS  & mIoU  & mAcc  \\ 
\hline 
wo $\mathcal{L}_{GR}$ & Ours (SH=3)     & \textbf{20.872} & \textbf{0.704} & \textbf{0.322} & \textbf{16.330} & \textbf{52.486} \\
\hline 
\multirow{4}{*}{w $\mathcal{L}_{GR}$} 
  & SH=0   & 19.040    & 0.674     & 0.372     & 15.066    & 45.537  \\
~ & SH=1   & 18.791    & 0.669     & 0.376     & 13.953    & 45.471  \\
~ & SH=2   & 18.780    & 0.668     & 0.378     & 13.664    & 44.995  \\
~ & SH=3   & 18.766    & 0.664     & 0.380     & 12.358    & 44.700  \\ 
\Xhline{3\arrayrulewidth}
\end{tabular}
\end{adjustbox}
\caption{
Results of different spherical harmonics parameters. 
} 
\label{table_sh}
\end{table}

\section{More Visualization Results} \label{More_Visualization_results}
\begin{figure*}
\centering
\includegraphics[width=\linewidth]{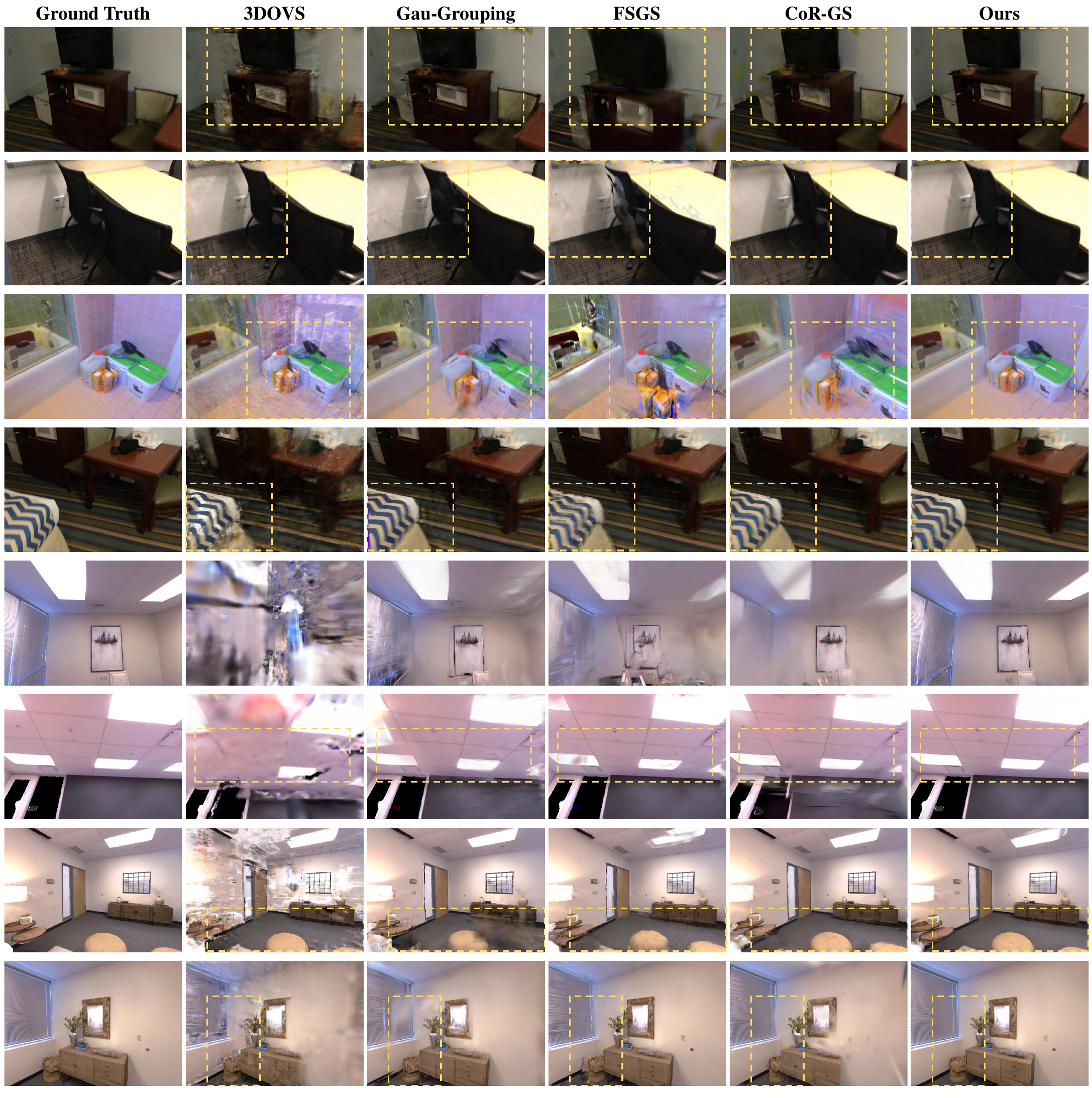}
\caption{
Visual reconstruction results on novel views from the ScanNet dataset (1$^{st}$ $\sim$ 4$^{th}$ Rows) and the Replica dataset (5$^{th}$ $\sim$ 8$^{th}$ Rows), using 12 input views for training. 
Our approach achieves superior global structure and photo-realistic details, attributed to our enhanced Gaussian representation and effective view-consistency constraints. 
More detailed analyses refer to Section \ref{More_Visualization_results}. 
}
\label{fig:vis_comparison_recon_supple}
\end{figure*}

\begin{figure*}
\centering
\includegraphics[width=.99\linewidth]{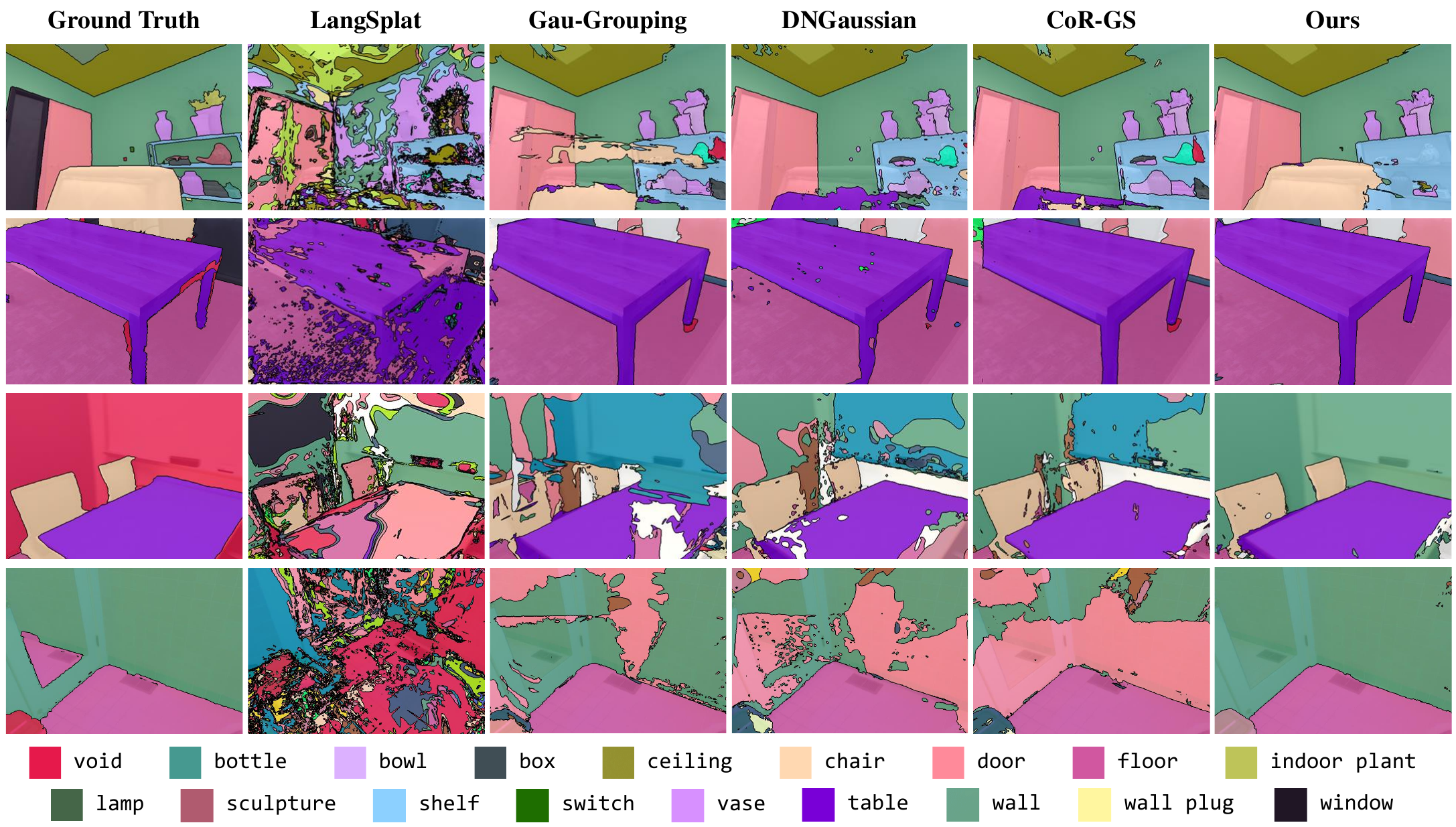}
\vspace{-3mm}
\caption{
Visual open-world segmentation results on novel views from the Replica dataset (1$^{st}$ Row) and the ScanNet dataset (3$^{rd}$ $\sim$ 4$^{th}$ Rows) when using the CLIP \cite{OpenAICLIP} to optimize Gaussian semantic attributes, with 12 training views.   
Our method produces more accurate and complete results thanks to the enhanced Gaussian points distribution and semantic consistency constraints. 
}
\vspace{-2mm}
\label{fig:vis_comparison_clip_supple}
\end{figure*}

\begin{figure*}
\centering
\includegraphics[width=.99\linewidth]{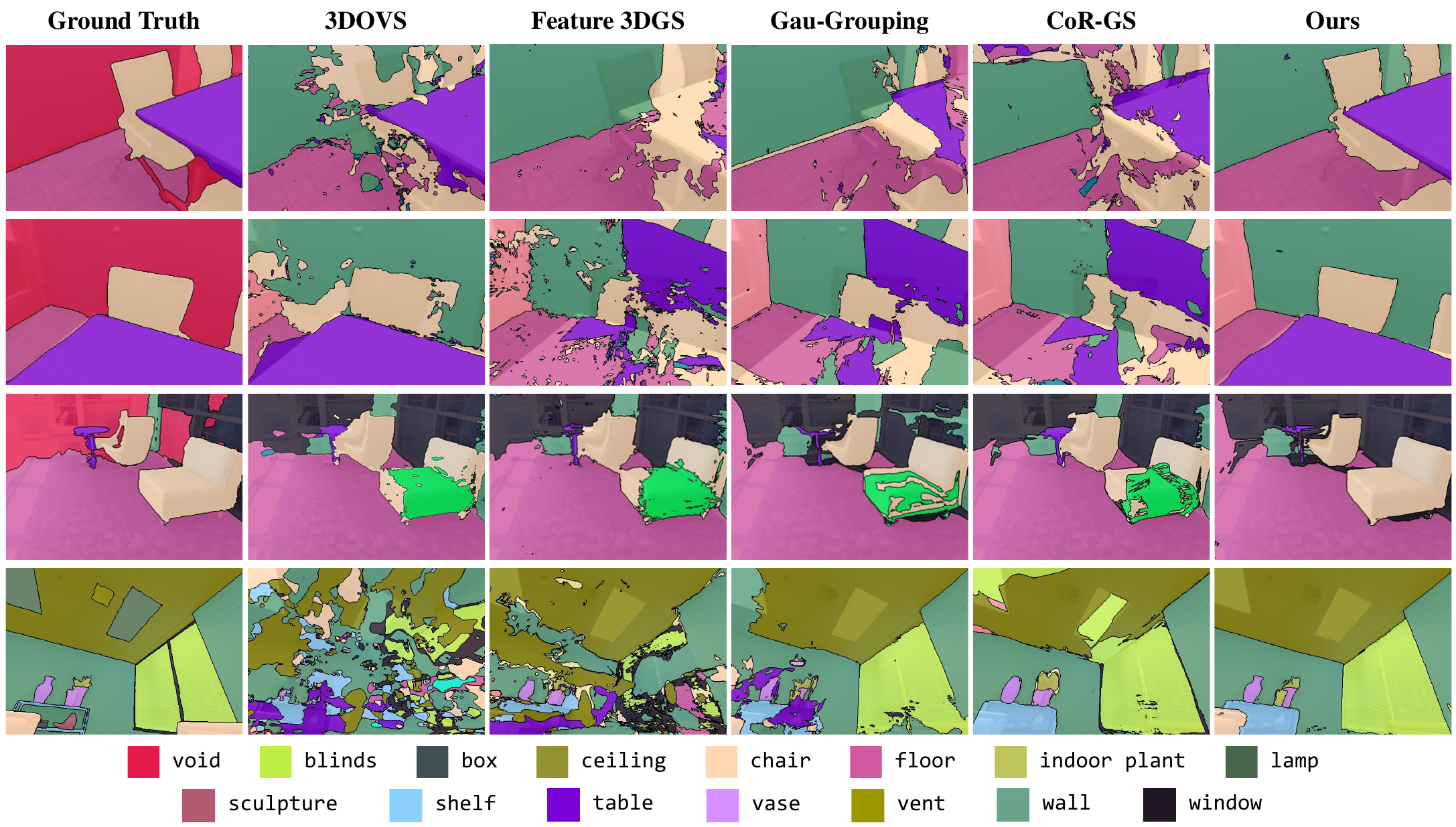}
\vspace{-3mm}
\caption{
Visual open-world segmentation results on novel views from the ScanNet dataset (1$^{st}$ $\sim$ 3$^{rd}$ Rows) and the Replica dataset (4$^{th}$ Row) when using the CLIP-LSeg model \cite{LSeg} for optimizing Gaussian semantic attributes, with 12 training views.  
Our approach consistently demonstrates superior precision. 
}
\label{fig:vis_comparison_lseg_supple}
\end{figure*}

We provide additional qualitative results in Fig. \ref{fig:vis_comparison_recon_supple}, \ref{fig:vis_comparison_clip_supple} and \ref{fig:vis_comparison_lseg_supple} to better illustrate the effectiveness of our method.  

\textit{i) Reconstruction.} 
As shown in Fig. \ref{fig:vis_comparison_recon_supple}, the results produced by 3DOVS and Gau-Grouping exhibit significant artifacts due to the limited radiance field representation quality when relying solely on sparse-input training data. 
Although FSGS and CoR-GS introduce additional geometric constraints to enhance Gaussian radiance fields under sparse input conditions, they struggle to obtain photo-realistic details. 
In contrast, our approach effectively reduces artifacts and ambiguity, presenting a more robust global structure with finer details, especially in rendered views distant from the training viewpoints (\eg 3$^{rd}$ and 5$^{th}$ rows). 
These improvements can be attributed to our SGI and SPC strategies, which contribute to an enhanced Gaussian distribution and enforce effective view-consistency supervision, thereby improving the overall quality of Gaussian representation.

\textit{ii) Open-world Segmentation.} 
As shown in Fig. \ref{fig:vis_comparison_clip_supple}, when using the CLIP model \cite{OpenAICLIP} for optimizing Gaussian semantic attributes, existing methods struggle to achieve precise object boundaries and maintain object integrity. 
Specifically, 3D open-vocabulary segmentation methods, such as LangSplat and Gau-Grouping, are hindered by the inferior Gaussian point distributions, which consequently impede the semantic Gaussian representation and easily lead to noisy rendered semantic results (\eg \textit{Chair} in the 1$^{st}$ row).  
Notably, LangSplat exhibits more pronounced noise (1$^{st}$ $\sim$ 2$^{nd}$ rows) and even inaccurate semantic renderings (3$^{rd}$ $\sim$ 4$^{th}$ rows) in sparse-input scenarios. This stems from a key factor: during semantic parameter optimization, LangSplat inherits Gaussian attributes (\ie position, scaling, and rotation) derived from sparse-input scene reconstruction using the vanilla 3DGS, and only optimizes the semantic parameter to obtain semantic Gaussian representation. 
By fixing the Gaussian primitives' attributes, LangSplat prevents flexible adjustment of semantic representations, thereby inheriting the inherent bottlenecks of sparse-input scene reconstruction and significantly compromising segmentation performance.  

Moreover, while sparse-input free-view synthesis methods, such as DNGaussian and CoR-GS, introduce geometric constraints to improve Gaussian representation, they still easily face challenges with semantic ambiguity due to insufficient semantic consistency supervision. 

In contrast, our method delivers robust and accurate segmentation results, benefiting from the enhanced Gaussian distribution and effective semantic-prompt consistency supervision. 
Additional visualizations in Fig. \ref{fig:vis_comparison_lseg_supple} further demonstrate that, when optimized with the CLIP-LSeg model \cite{LSeg}, our approach consistently outperforms others across diverse scenes, underscoring its robustness in 3D open-vocabulary semantic understanding.

In summary, our approach simultaneously delivers photorealistic rendering quality and superior segmentation performance across diverse scenes.  
This comprehensive evaluation highlights the effectiveness of our approach for sparse-input open-world free-view synthesis.

\section{Per-scene Qualitative Results} \label{Scene_Qualitative_results}
In Tables \ref{table:SOTA_clip_per_scene} and \ref{table:SOTA_lseg_per_scene}, we present per-scene quantitative comparisons of reconstruction and segmentation results on novel views, utilizing {CLIP} and {CLIP-LSeg} for optimizing Gaussian semantic attributes, respectively. 
Overall, our approach consistently surpasses other state-of-the-art methods in terms of reconstruction and segmentation on synthetic and real-world scenes. 
These results highlight the effectiveness of our method in indoor open-world free-view synthesis using sparse-input data.

\begin{table*}[!t]
\renewcommand{\arraystretch}{1.1}
\centering
\begin{adjustbox} {width=\linewidth}
\begin{tabular}{ c | p{1.3cm}<{\centering} p{1.3cm}<{\centering} p{1.3cm}<{\centering} p{1.3cm}<{\centering} p{1.3cm}<{\centering} | p{1.3cm}<{\centering} p{1.3cm}<{\centering} p{1.3cm}<{\centering} p{1.3cm}<{\centering} p{1.3cm}<{\centering} p{1.3cm}<{\centering} }
\Xhline{3\arrayrulewidth}
\multirow{2}{*}{Method} & \multicolumn{5}{c|}{Room0} & \multicolumn{5}{c}{Room1}\\ 
& PSNR $\uparrow$ & SSIM $\uparrow$ & LPIPS $\downarrow$ & mIoU $\uparrow$ & mAcc $\uparrow$ 
& PSNR $\uparrow$ & SSIM $\uparrow$ & LPIPS $\downarrow$ & mIoU $\uparrow$ & mAcc $\uparrow$ \\ 
\hline 
3DOVS        \cite{3DOVS}       & 16.198 & 0.554 & 0.467 & 5.547  & 32.699 & 14.643 & 0.496 & 0.674 & 3.166  & 18.421 \\
Feature 3DGS \cite{feature3dgs} & 15.991 & 0.595 & 0.425 & 6.402 & 32.734 & 17.485 & 0.674 & 0.373 & 5.959  & 22.927 \\
LangSplat    \cite{langsplat}   & 16.481 & 0.580 & 0.407 & 2.924  & 20.102 & 17.116 & 0.639 & 0.393 & 2.167  & 17.038 \\
Gau-Grouping \cite{gaugrouping} & 15.978 & 0.599 & 0.436 & 10.712 & 40.774 & 16.400 & 0.639 & 0.410 & 12.709 & 46.494 \\
DNGaussian   \cite{dngaussian}  & 16.005 & 0.601 & 0.492 & 11.274 & 42.377 & 19.353 & 0.691 & 0.406 & 16.883 & 53.758 \\
FSGS \cite{fsgs}                & 18.437 & 0.650 & 0.365 & 12.695 & 44.934 & 18.516 & 0.671 & 0.356 & 16.065 & 56.217 \\
CoR-GS \cite{CoRGS}             & 18.813 & 0.688 & 0.357 & 12.747 & 44.289 & 19.526 & 0.725 & 0.327 & 16.158 & 51.927 \\
Ours & \textbf{20.872} & \textbf{0.704} & \textbf{0.322} & \textbf{16.330} & \textbf{52.486} & \textbf{22.363} & \textbf{0.750} & \textbf{0.264} & \textbf{29.022} & \textbf{74.166} \\
\hline\hline
\multirow{2}{*}{Method} & \multicolumn{5}{c|}{Room2} & \multicolumn{5}{c}{Office0}\\ 
& PSNR $\uparrow$ & SSIM $\uparrow$ & LPIPS $\downarrow$ & mIoU $\uparrow$ & mAcc $\uparrow$ 
& PSNR $\uparrow$ & SSIM $\uparrow$ & LPIPS $\downarrow$ & mIoU $\uparrow$ & mAcc $\uparrow$ \\ 
\hline
3DOVS        \cite{3DOVS}       & 13.536 & 0.567 & 0.669 & 4.633  & 25.459 & 14.317 & 0.485 & 0.680 & 2.156  & 9.580  \\
Feature 3DGS \cite{feature3dgs} & 15.484 & 0.705 & 0.384 & 7.139 & 29.271 & 20.719 & 0.768 & 0.279 & 6.430  & 21.632 \\
LangSplat    \cite{langsplat}   & 18.092 & 0.740 & 0.314 & 4.277  & 27.811 & 21.543 & 0.774 & 0.277 & 1.852  & 11.097 \\
Gau-Grouping \cite{gaugrouping} & 19.359 & 0.756 & 0.313 & 17.657 & 69.003 & 20.128 & 0.738 & 0.334 & 11.485 & 29.045 \\
DNGaussian   \cite{dngaussian}  & 20.626 & 0.767 & 0.353 & 17.229 & 70.091 & 20.552 & 0.759 & 0.362 & 13.055 & 32.788 \\
FSGS \cite{fsgs}                & 18.295 & 0.740 & 0.343 & 15.562 & 65.899 & 21.762 & 0.789 & 0.262 & 13.548 & 31.720 \\
CoR-GS \cite{CoRGS}             & 18.648 & 0.783 & 0.303 & 15.270 & 64.398 & 22.056 & 0.804 & 0.251 & 13.020 & 31.989 \\
Ours & \textbf{23.267} & \textbf{0.835} & \textbf{0.209} & \textbf{21.112} & \textbf{73.906} & \textbf{23.382} & \textbf{0.821} & \textbf{0.246} & \textbf{16.488} & \textbf{34.438} \\
\hline\hline
\multirow{2}{*}{Method} & \multicolumn{5}{c|}{Office2} & \multicolumn{5}{c}{Office4}\\ 
& PSNR $\uparrow$ & SSIM $\uparrow$ & LPIPS $\downarrow$ & mIoU $\uparrow$ & mAcc $\uparrow$ 
& PSNR $\uparrow$ & SSIM $\uparrow$ & LPIPS $\downarrow$ & mIoU $\uparrow$ & mAcc $\uparrow$ \\ 
\hline
3DOVS        \cite{3DOVS}       & 9.716  & 0.496 & 0.760 & 3.228 & 18.629 & 10.789 & 0.490 & 0.726 & 4.748  & 23.844 \\
Feature 3DGS \cite{feature3dgs} & 16.758 & 0.752 & 0.338 & 8.096 & 34.719 & 15.915 & 0.708 & 0.352 & 10.416 & 39.970 \\
LangSplat    \cite{langsplat}   & 18.237 & 0.779 & 0.779 & 3.802  & 12.832 & 14.821 & 0.694 & 0.348 & 4.729  & 22.844 \\
Gau-Grouping \cite{gaugrouping} & 16.133 & 0.716 & 0.381 & 19.108 & 67.423 & 15.784 & 0.695 & 0.380 & 18.500 & 49.792 \\
DNGaussian   \cite{dngaussian}  & 19.100 & 0.765 & 0.369 & 21.779 & 76.004 & 15.438 & 0.686 & 0.429 & 19.723 & 51.380 \\
FSGS \cite{fsgs}                & 18.261 & 0.777 & 0.310 & 19.737 & 71.854 & 16.893 & 0.724 & 0.334 & 18.998 & 50.224 \\
CoR-GS \cite{CoRGS}             & 17.313 & 0.781 & 0.325 & 18.187 & 70.760 & 17.374 & 0.768 & 0.309 & 20.220 & 51.490 \\
Ours & \textbf{23.095} & \textbf{0.858} & \textbf{0.204} & \textbf{35.109} & \textbf{87.375} & \textbf{19.087} & \textbf{0.785} & \textbf{0.275} & \textbf{25.697} & \textbf{57.201} \\
\hline\hline
\multirow{2}{*}{Method} & \multicolumn{5}{c|}{scene0004} & \multicolumn{5}{c}{scene0389}\\ 
& PSNR $\uparrow$ & SSIM $\uparrow$ & LPIPS $\downarrow$ & mIoU $\uparrow$ & mAcc $\uparrow$ 
& PSNR $\uparrow$ & SSIM $\uparrow$ & LPIPS $\downarrow$ & mIoU $\uparrow$ & mAcc $\uparrow$ \\ 
\hline
3DOVS        \cite{3DOVS}       & 17.894 & 0.615 & 0.496 & 10.406 & 51.940 & 21.550 & 0.731 & 0.525 & 7.919  & 45.842 \\
Feature 3DGS \cite{feature3dgs} & 16.489 & 0.600 & 0.434 & 8.447 & 51.692 & 21.061 & 0.741 & 0.383 & 10.063 & 46.725 \\
LangSplat    \cite{langsplat}   & 16.656 & 0.605 & 0.427 & 5.499  & 43.347 & 23.720 & 0.791 & 0.352 & 4.545  & 34.339 \\
Gau-Grouping \cite{gaugrouping} & 17.837 & 0.634 & 0.424 & 10.416 & 54.014 & 23.241 & 0.785 & 0.366 & 24.414 & 78.311 \\
DNGaussian   \cite{dngaussian}  & 17.810 & 0.670 & 0.506 & 9.460  & 53.999 & 20.818 & 0.739 & 0.442 & 20.590 & 70.258 \\
FSGS \cite{fsgs}                & 19.011 & 0.645 & 0.419 & 9.934 & 55.340 & 24.681 & 0.796 & 0.354 & 21.265 & 73.597 \\ 
CoR-GS \cite{CoRGS}             & 18.961 & 0.662 & 0.407 & 10.582 & 54.445 & 25.098 & 0.812 & 0.338 & 23.462 & 75.366 \\
Ours & \textbf{19.131} & \textbf{0.677} & \textbf{0.404} & \textbf{13.411} & \textbf{64.718} & \textbf{27.232} & \textbf{0.838} & \textbf{0.329} & \textbf{48.243} & \textbf{82.017} \\
\hline\hline
\multirow{2}{*}{Method} & \multicolumn{5}{c|}{scene0494} & \multicolumn{5}{c}{scene0693}\\ 
& PSNR $\uparrow$ & SSIM $\uparrow$ & LPIPS $\downarrow$ & mIoU $\uparrow$ & mAcc $\uparrow$ 
& PSNR $\uparrow$ & SSIM $\uparrow$ & LPIPS $\downarrow$ & mIoU $\uparrow$ & mAcc $\uparrow$ \\ 
\hline
3DOVS        \cite{3DOVS}       & 16.446 & 0.663 & 0.577 & 6.162  & 31.342 & 18.996 & 0.561 & 0.580 & 6.516  & 16.300 \\
Feature 3DGS \cite{feature3dgs} & 14.304 & 0.649 & 0.445 & 5.571 & 31.384 & 20.497 & 0.635 & 0.436 & 5.309  & 22.209 \\
LangSplat    \cite{langsplat}   & 15.235 & 0.646 & 0.646 & 2.409  & 15.807 & 19.511 & 0.610 & 0.479 & 0.903  & 4.469  \\
Gau-Grouping \cite{gaugrouping} & 15.311 & 0.648 & 0.447 & 10.529 & 46.358 & 19.633 & 0.635 & 0.446 & 15.330 & 50.737 \\
DNGaussian   \cite{dngaussian}  & 15.820 & 0.699 & 0.473 & 12.564 & 50.876 & 19.120 & 0.656 & 0.516 & 13.036 & 44.122 \\
FSGS \cite{fsgs}                & 16.511 & 0.659 & 0.433 & 11.857 & 53.064 & 22.094 & 0.682 & 0.417 & 13.069 & 45.399 \\ 
CoR-GS \cite{CoRGS}             & 16.303 & 0.706 & 0.416 & 11.175 & 49.935 & 22.897 & 0.703 & 0.410 & 12.628 & 43.501 \\
Ours & \textbf{18.244} & \textbf{0.725} & \textbf{0.386} & \textbf{26.651} & \textbf{67.365} & \textbf{24.998} & \textbf{0.723} & \textbf{0.354} & \textbf{27.472} & \textbf{61.194} \\ 
\Xhline{3\arrayrulewidth}
\end{tabular}
\end{adjustbox}
\caption{Quantitative results of reconstruction and segmentation on novel views across various scenes from Replica and ScanNet datasets, using the CLIP \cite{OpenAICLIP} for optimizing Gaussian semantic attributes with 12 training views. 
Our approach achieves \textbf{superior performances} in both reconstruction quality and segmentation accuracy. 
} 
\label{table:SOTA_clip_per_scene}
\end{table*}

\begin{table*}[!t]
\renewcommand{\arraystretch}{1.2}
\centering
\begin{adjustbox} {width=\linewidth}
\begin{tabular}{ c | p{1.3cm}<{\centering} p{1.3cm}<{\centering} p{1.3cm}<{\centering} p{1.3cm}<{\centering} p{1.3cm}<{\centering} | p{1.3cm}<{\centering} p{1.3cm}<{\centering} p{1.3cm}<{\centering} p{1.3cm}<{\centering} p{1.3cm}<{\centering} p{1.3cm}<{\centering} }
\Xhline{3\arrayrulewidth}
\multirow{2}{*}{Method} & \multicolumn{5}{c|}{Room0} & \multicolumn{5}{c}{Room1}\\ 
& PSNR $\uparrow$ & SSIM $\uparrow$ & LPIPS $\downarrow$ & mIoU $\uparrow$ & mAcc $\uparrow$ 
& PSNR $\uparrow$ & SSIM $\uparrow$ & LPIPS $\downarrow$ & mIoU $\uparrow$ & mAcc $\uparrow$ \\ 
\hline 
3DOVS        \cite{3DOVS}       & 16.277 & 0.557 & 0.464 & 17.789 & 62.604 & 14.717 & 0.513 & 0.661 & 7.771 & 41.968 \\ 
Feature 3DGS \cite{feature3dgs} & 16.992 & 0.599 & 0.405 & 14.700 & 58.249 & 17.532 & 0.650 & 0.373 & 12.257 & 57.894 \\
Gau-Grouping \cite{gaugrouping} & 17.548 & 0.636 & 0.392 & 20.618 & 65.525 & 18.857 & 0.681 & 0.348 & 16.899 & 63.053 \\
DNGaussian   \cite{dngaussian}  & 17.168 & 0.633 & 0.467 & 19.914 & 62.975 & 20.731 & 0.722 & 0.378 & 21.399 & 68.646 \\
FSGS \cite{fsgs}                & 19.720 & 0.696 & 0.336 & 21.967 & 66.214 & 20.969 & 0.731 & 0.303 & 19.626 & 67.325 \\
CoR-GS \cite{CoRGS}             & 19.362 & 0.707 & 0.351 & 19.377 & 61.534 & 19.603 & 0.732 & 0.329 & 16.332 & 60.915 \\
Ours & \textbf{20.944} & \textbf{0.716} & \textbf{0.311} & \textbf{25.740} & \textbf{70.296} & \textbf{22.541} & \textbf{0.761} & \textbf{0.252} & \textbf{23.539} & \textbf{71.692} \\
\hline\hline
\multirow{2}{*}{Method} & \multicolumn{5}{c|}{Room2} & \multicolumn{5}{c}{Office0}\\ 
& PSNR $\uparrow$ & SSIM $\uparrow$ & LPIPS $\downarrow$ & mIoU $\uparrow$ & mAcc $\uparrow$ 
& PSNR $\uparrow$ & SSIM $\uparrow$ & LPIPS $\downarrow$ & mIoU $\uparrow$ & mAcc $\uparrow$ \\ 
\hline
3DOVS        \cite{3DOVS}       & 13.540 & 0.570 & 0.660 & 7.886  & 44.913 & 14.580 & 0.483 & 0.677 & 4.347 & 25.848 \\
Feature 3DGS \cite{feature3dgs} & 18.338 & 0.742 & 0.342 & 16.536 & 67.860 & 21.738 & 0.771 & 0.285 & 13.247 & 56.463 \\
Gau-Grouping \cite{gaugrouping} & 17.910 & 0.737 & 0.341 & 18.196 & 67.612 & 20.530 & 0.751 & 0.313 & 16.006 & 54.395 \\
DNGaussian   \cite{dngaussian}  & 20.991 & 0.784 & 0.333 & 24.145 & 76.987 & 21.700 & 0.785 & 0.352 & 18.991 & 59.040 \\ 
FSGS \cite{fsgs}                & 20.149 & 0.781 & 0.272 & 24.636 & 76.413 & 22.580 & 0.810 & 0.252 & 19.137 & 59.542 \\
CoR-GS \cite{CoRGS}             & 19.893 & 0.805 & 0.265 & 23.611 & 74.854 & 22.715 & 0.808 & 0.253 & 19.845 & 59.545 \\ 
Ours & \textbf{22.879} & \textbf{0.838} & \textbf{0.210} & \textbf{26.995} & \textbf{78.006} & \textbf{23.134} & \textbf{0.819} & \textbf{0.249} & \textbf{23.166} & \textbf{62.666} \\
\hline\hline
\multirow{2}{*}{Method} & \multicolumn{5}{c|}{Office2} & \multicolumn{5}{c}{Office4}\\ 
& PSNR $\uparrow$ & SSIM $\uparrow$ & LPIPS $\downarrow$ & mIoU $\uparrow$ & mAcc $\uparrow$ 
& PSNR $\uparrow$ & SSIM $\uparrow$ & LPIPS $\downarrow$ & mIoU $\uparrow$ & mAcc $\uparrow$ \\ 
\hline
3DOVS        \cite{3DOVS}       & 9.687  & 0.497 & 0.753 & 6.912  & 40.524 & 11.157 & 0.502 & 0.721 & 8.448 & 35.599 \\
Feature 3DGS \cite{feature3dgs} & 18.812 & 0.788 & 0.281 & 20.073 & 79.211 & 15.513 & 0.701 & 0.346 & 21.328 & 70.054 \\
Gau-Grouping \cite{gaugrouping} & 17.130 & 0.756 & 0.323 & 26.287 & 78.060 & 14.748 & 0.692 & 0.384 & 23.505 & 70.468 \\
DNGaussian   \cite{dngaussian}  & 21.082 & 0.822 & 0.311 & 27.715 & 81.715 & 18.115 & 0.748 & 0.381 & 29.989 & 74.869 \\
FSGS \cite{fsgs}                & 20.486 & 0.817 & 0.251 & 27.564 & 82.127 & 18.324 & 0.772 & 0.294 & 29.397 & 75.789 \\ 
CoR-GS \cite{CoRGS}             & 20.354 & 0.830 & 0.255 & 26.757 & 82.480 & 18.467 & 0.792 & 0.286 & 29.053 & 74.079 \\
Ours & \textbf{23.413} & \textbf{0.868} & \textbf{0.200} & \textbf{39.293} & \textbf{88.195} & \textbf{19.825} & \textbf{0.798} & \textbf{0.266} & \textbf{36.304} & \textbf{82.039} \\
\hline\hline
\multirow{2}{*}{Method} & \multicolumn{5}{c|}{scene0004} & \multicolumn{5}{c}{scene0389}\\ 
& PSNR $\uparrow$ & SSIM $\uparrow$ & LPIPS $\downarrow$ & mIoU $\uparrow$ & mAcc $\uparrow$ 
& PSNR $\uparrow$ & SSIM $\uparrow$ & LPIPS $\downarrow$ & mIoU $\uparrow$ & mAcc $\uparrow$ \\ 
\hline
3DOVS        \cite{3DOVS}       & 17.829 & 0.616 & 0.498 & 23.079 & 75.464 & 21.882 & 0.739 & 0.515 & 24.401 & 81.412 \\
Feature 3DGS \cite{feature3dgs} & 18.038 & 0.618 & 0.406 & 16.215 & 77.017 & 24.682 & 0.806 & 0.337 & 31.594 & 89.220 \\
Gau-Grouping \cite{gaugrouping} & 17.975 & 0.631 & 0.411 & 28.301 & 74.673 & 22.931 & 0.779 & 0.365 & 43.316 & 83.081 \\
DNGaussian   \cite{dngaussian}  & 17.853 & 0.636 & 0.487 & 25.833 & 73.883 & 25.363 & 0.819 & 0.374 & 39.106 & 89.642 \\
FSGS \cite{fsgs}                & 19.399 & 0.672 & 0.411 & 28.743 & 78.000 & 26.784 & 0.844 & 0.328 & 39.576 & 89.337 \\
CoR-GS \cite{CoRGS}             & 19.331 & 0.673 & 0.412 & 29.863 & 79.074 & 26.573 & 0.828 & 0.333 & 37.277 & 89.190 \\
Ours  & \textbf{19.631} & \textbf{0.681} & \textbf{0.403} & \textbf{40.120} & \textbf{84.554} & \textbf{27.512} & \textbf{0.847} & \textbf{0.320} & \textbf{63.541} & \textbf{89.646} \\
\hline\hline
\multirow{2}{*}{Method} & \multicolumn{5}{c|}{scene0494} & \multicolumn{5}{c}{scene0693}\\ 
& PSNR $\uparrow$ & SSIM $\uparrow$ & LPIPS $\downarrow$ & mIoU $\uparrow$ & mAcc $\uparrow$ 
& PSNR $\uparrow$ & SSIM $\uparrow$ & LPIPS $\downarrow$ & mIoU $\uparrow$ & mAcc $\uparrow$ \\ 
\hline
3DOVS        \cite{3DOVS}       & 16.703 & 0.676 & 0.560 & 26.038 & 75.165 & 18.930 & 0.577 & 0.566 & 22.487 & 59.412 \\
Feature 3DGS \cite{feature3dgs} & 15.448 & 0.659 & 0.426 & 17.940 & 67.083 & 20.641 & 0.629 & 0.443 & 18.461 & 57.569 \\
Gau-Grouping \cite{gaugrouping} & 15.594 & 0.670 & 0.428 & 34.902 & 67.838 & 20.255 & 0.649 & 0.471 & 27.678 & 68.902 \\
DNGaussian   \cite{dngaussian}  & 15.240 & 0.664 & 0.475 & 32.891 & 67.215 & 22.469 & 0.704 & 0.468 & 30.657 & 68.838 \\
FSGS \cite{fsgs}                & 17.058 & 0.691 & 0.413 & 37.701 & 74.877 & 24.257 & 0.713 & 0.391 & 32.422 & 69.146 \\ 
CoR-GS \cite{CoRGS}             & 17.386 & 0.716 & 0.406 & 38.263 & 75.702 & 23.913 & 0.724 & 0.399 & 31.276 & 67.810  \\
Ours & \textbf{19.711} & \textbf{0.756} & \textbf{0.355} & \textbf{52.016} & \textbf{81.288} & \textbf{25.315} & \textbf{0.737} & \textbf{0.357} & \textbf{45.407} & \textbf{78.849} \\ 
\Xhline{3\arrayrulewidth}
\end{tabular}
\end{adjustbox}
\caption{Quantitative results of reconstruction and segmentation on novel views across various scenes from Replica and ScanNet datasets, using CLIP-LSeg \cite{LSeg} for optimizing Gaussian semantic attributes with 12 training views. 
Our approach achieves \textbf{superior performances} in both reconstruction quality and segmentation accuracy. 
} 
\label{table:SOTA_lseg_per_scene}
\end{table*}

\end{document}